%% file: template.tex
\documentclass{article}
\usepackage[final]{colm2026_conference}

\usepackage{microtype}
\usepackage{hyperref}
\usepackage{url}
\usepackage{booktabs}
\usepackage{graphicx}
\usepackage{subcaption}
\usepackage{multirow}
\usepackage{xcolor}
\usepackage{colortbl}
\usepackage{enumitem}
\usepackage{amsmath,amsfonts,amssymb,bm}
\usepackage{algorithm}
\usepackage{algpseudocode}
\usepackage{xspace}
\usepackage{tabularx}
\usepackage{makecell}
\usepackage{adjustbox}
\usepackage{float}

\input{math_commands}

\definecolor{darkblue}{rgb}{0, 0, 0.5}
\hypersetup{colorlinks=true, citecolor=darkblue, linkcolor=darkblue, urlcolor=darkblue}

\newcommand{\gc}[1]{\textcolor{black!50}{#1}}

\newcommand{\skillrouter}{\textsc{SkillRouter}\xspace}

\newcommand{\skill}{s}
\newcommand{\skillset}{\mathcal{S}}
\newcommand{\query}{q}
\newcommand{\gtset}{\mathcal{G}}

\title{SkillRouter: Skill Routing for LLM Agents at Scale}

\author{
  \textbf{YanZhao Zheng, ZhenTao Zhang, Chao Ma, YuanQiang Yu, JiHuai Zhu, Yong Wu}\\[-1pt]
  \textbf{Tianze Xu, Baohua Dong\thanks{Corresponding author.}, Hangcheng Zhu, Ruohui Huang, Gang Yu}\\[2pt]
  Alibaba Group, Hangzhou, China \\
  {\scriptsize\texttt{zhengyanzhao.zyz, zhangzhentao.zzt, mc524716, yuyuanqiang.yyq, zhujihuai.zjh, wy517954,}}\\[-1pt]
  {\scriptsize\texttt{xutianze.xtz, baohua.dbh, linran.lr09, wentong, ruohai}@alibaba-inc.com}\vspace{-1em}
}

\begin{document}

\maketitle
\fancyhead{}
\renewcommand{\headrulewidth}{0pt}

\input{sections/abstract}

\input{sections/introduction}

\input{sections/problem}

\input{sections/body_study}

\input{sections/method}

\input{sections/experiments}

\input{sections/related_work}

\input{sections/conclusion}

\bibliographystyle{colm2026_conference}
\bibliography{references}

\clearpage
\appendix
\raggedbottom
\input{sections/appendix}

\end{document}

%% file: math_commands.tex
\usepackage{amsfonts,bm}

\def\eqref#1{equation~\ref{#1}}

\def\1{\bm{1}}

\DeclareMathAlphabet{\mathsfit}{\encodingdefault}{\sfdefault}{m}{sl}
\SetMathAlphabet{\mathsfit}{bold}{\encodingdefault}{\sfdefault}{bx}{n}



%% file: sections/abstract.tex
\begin{abstract}
Reusable skills let LLM agents package task-specific procedures, tool affordances, and execution guidance into modular building blocks. As skill ecosystems grow to tens of thousands of entries, exposing every skill at inference time becomes infeasible.
This creates a skill-routing problem: given a user task, the system must identify relevant skills before downstream planning or execution.
Existing agent stacks often rely on progressive disclosure, exposing only skill names and descriptions while hiding the full implementation body.
We examine this design choice on a SkillsBench-derived benchmark with approximately 80K candidate skills, targeting the practically important setting of large skill registries with heavy overlap.
Across representative dense and reranking baselines on this setting, hiding the skill body causes a 37--44 percentage point drop in routing accuracy.
Stronger controls show that the missing signal is body-resident rather than a simple length artifact: body-distilled descriptions recover part of the gap, but remain 7--21 points below direct all-field routing, while a metadata-only encoder trained with the same data remains 14.0 points below its all-field counterpart.
Motivated by this finding, we present \skillrouter, a compact 1.2B body-aware retrieve-and-rerank pipeline.
\skillrouter achieves 74.0\% Hit@1 on our benchmark---the strongest average top-1 routing performance among the baselines we evaluate---while using 13$\times$ fewer parameters and running 5.8$\times$ faster than the strongest base pipeline.
The ranking gains further generalize to a supplementary benchmark independently constructed from three skill sources.
In a complementary end-to-end study across four coding agents, routing gains transfer to improved task success, with larger gains for more capable agents.
\end{abstract}

%% file: sections/introduction.tex
\section{Introduction}
\label{sec:intro}

Skills have emerged as a practical abstraction for extending LLM agents with reusable procedures, tool knowledge, and execution guidance. Recent coding-agent products such as Claude Code, Codex, and OpenClaw expose reusable skills as a first-class capability \citep{anthropic2025claude,openai2025codex,openclaw2026skills}. These systems reflect the growing use of skill registries in real deployments.
Presenting every skill to the agent is infeasible, so real systems need \textit{skill routing}: retrieving the right skill from a large pool given a user task.
This setting has an important asymmetry: the routing component can inspect all skill fields, while the agent that eventually consumes the skill usually sees only its name and description.
In deployed agent stacks, this upstream routing decision is a high-leverage bottleneck: once the wrong skill shortlist is surfaced, downstream planning and execution have little chance to recover.
The question is therefore not only whether an agent can \textit{use} a provided skill, but whether the system can \textit{find} the right skill under severe pool-scale confusion.

Current agent frameworks implicitly treat metadata as sufficient for selection, yet this assumption has not been tested under open-registry scale and overlap.
Existing benchmarks such as SkillsBench \citep{skillsbench2026}, ToolBench \citep{qin2023toolllm}, and MetaTool \citep{huang2024metatool} study downstream tool use or tool-choice behavior, but they do not directly evaluate large-pool upstream skill routing under hidden implementations.
On the retrieval side, ToolRet studies retrieval over 43K heterogeneous tools \citep{shi2025toolret}, while ToolRerank, CRAFT, and AgentSkillOS study adaptive reranking, toolset synthesis, and skill organization/orchestration, respectively \citep{xu2024toolrerank,yuan2024craft,li2026organizing}.
These works establish large-scale tool and skill retrieval as important, but do not isolate which field drives routing under the progressive-disclosure asymmetry studied here.
Our goal is not to claim that every skill-routing benchmark exhibits the same failure mode, but to study the practically important setting of large skill registries with heavy overlap, where many candidates can appear relevant for the same query.

We study skill routing on a benchmark with ${\sim}$80K skills and 75 expert-verified SkillsBench-derived queries that instantiate this setting.
Our central empirical finding is that, on this setting, raw name-and-description metadata omits critical body-resident routing signal: removing the body causes 37--44pp drops across representative dense and reranking baselines.
Length-controlled attention, description-length stratification, body-distilled descriptions, and a metadata-only fine-tuning control jointly rule out simple text-length and training-signal explanations.
Motivated by this observation, we build \skillrouter, a compact 1.2B body-aware retrieve-and-rerank pipeline.
The primary 1.2B configuration (0.6B encoder + 0.6B reranker) reaches 74.0\% Hit@1 and 70.4\% R@10, compared with 68.0\% Hit@1 for the strongest 16B base pipeline---achieving comparable or higher accuracy at 13$\times$ fewer parameters and 5.8$\times$ lower serving latency.
A scaled configuration with an 8B encoder and an 8B reranker (16B total) reaches 76.0\%.
We also validate transfer beyond retrieval metrics: in a complementary end-to-end study using the natural pool across four coding agents, \skillrouter improves average task success over the strongest base router in both top-1 and top-10 settings, with the benefit being more pronounced for more capable agents.
These downstream results should be read as end-to-end utility measurements rather than direct proxies for exhaustive gold-skill recovery, since the agent consumes a bounded shortlist rather than the abstract annotated set.
On a real-pool GPU benchmark the 1.2B pipeline serves queries at sub-second median latency.
We also evaluate the same checkpoints on SkillBench-Supp, a separate 256-query benchmark with LLM-generated queries and ground-truth skills drawn from three sources, providing a cross-source transfer test beyond the 75-query expert-verified core benchmark.

Our contributions are threefold:
\begin{enumerate}[nosep,leftmargin=*]
\item On two complementary benchmarks over an ${\sim}$80K-skill pool, we evaluate a 75-query expert-verified core set with Easy/Hard robustness tiers and a 256-query LLM-generated supplementary set independently constructed from three skill sources. We show that raw metadata omits critical body-resident routing signal (Section~\ref{sec:body_study}). Body-distilled metadata and matched metadata-only fine-tuning recover part, but not all, of the resulting gap.
\item We present \skillrouter, a compact body-aware retrieve-and-rerank pipeline built from standard IR components, and identify two training adaptations that are specifically necessary in homogeneous skill pools: false-negative filtering to handle near-duplicate skills, and listwise reranking loss to resolve fine-grained candidate competition.
\item We show that the routing gains transfer to a complementary end-to-end study using the natural pool across four coding agents, and we characterize the compact pipeline's efficiency--accuracy tradeoff on a real-pool GPU serving benchmark.
\end{enumerate}

%% file: sections/problem.tex
\section{Problem definition and benchmark}
\label{sec:problem}

\paragraph{Task and metrics.}
\label{sec:benchmark}
We study \textit{skill routing}: given a task query $\query$ and a large skill pool $\skillset = \{\skill_1,\ldots,\skill_N\}$, retrieve the skill set $\gtset_\query \subseteq \skillset$ needed to solve the task.
Each skill contains a \textit{name}, \textit{description}, and full implementation \textit{body}.
This creates a \textit{hidden-body asymmetry}: the routing system can inspect all skill fields, while the downstream agent initially sees only metadata.
We report Hit@1 as the primary top-1 routing metric, together with MRR@10, Recall@$K$ ($K\in\{10,20,50\}$; average fraction of ground-truth skills recovered), and FC@10 (fraction of queries whose full ground-truth skill set appears in the top 10).
For multi-skill queries, Hit@1 is defined mechanically as whether any required skill is ranked first. We therefore report Recall@$K$ and FC@10 to characterize shortlist and full-set coverage more directly.

\paragraph{Benchmark construction.}
\label{sec:bench_construct}
We build the benchmark from SkillsBench \citep{skillsbench2026}, which provides expert-curated task--skill mappings.
Starting from 87 SkillsBench tasks, we exclude 12 generic-only cases whose labels contain only file-type skills (e.g., \texttt{pdf} or \texttt{xlsx}) and retain \textbf{75 core queries}: \textbf{24 single-skill} and \textbf{51 multi-skill}.
We evaluate against an ${\sim}$80K-skill pool assembled from SkillsBench skills plus a large open-source skill collection spanning 51 categories, drawn from Claude Skill Registry Core \citep{majiayu2026claudeskillregistry}.
To probe robustness, we report two tiers: \textbf{Easy} with 78,361 candidate skills, and \textbf{Hard} with 79,141 candidates after adding \textbf{780} LLM-generated distractor skills that are topically related but functionally distinct.
All main results average Easy and Hard. The Hard additions are generated with three strategies: same domain but different problem, same technology but different use, and an over-generalized capability. They serve as a controlled stress test rather than an estimate of synthetic-skill prevalence.
Appendix~\ref{app:eval_details} reports the exact core-query selection protocol and a metadata audit of the pool, while Appendix~\ref{app:benchmark_data} details distractor generation.

\begin{table}[t]
\centering
\begin{adjustbox}{max width=\columnwidth}
\begin{tabular}{@{}p{2.5cm}p{3.5cm}p{7.0cm}@{}}
\toprule
\textbf{Type} & \textbf{Name} & \textbf{Description} \\
\midrule
Ground truth & \texttt{speech-to-text} & Transcribe audio/video locally with Whisper and return timestamped text. \\
Pool distractor & \texttt{audio-transcriber} & General-purpose cloud transcription service for uploaded audio files. \\
Hard distractor & \texttt{video-subtitle-sync} & Synchronize subtitle timing to video playback using audio cues. \\
\bottomrule
\end{tabular}
\end{adjustbox}
\caption{Illustrative benchmark example. Hard distractors remain topically plausible but fail the required function.}
\label{tab:benchmark_example}
\end{table}

\paragraph{Benchmark credibility and scope.}
SkillsBench provides expert-curated task--skill mappings rather than weakly inferred labels.
The Easy/Hard split isolates two failure modes: standard large-pool retrieval in Easy, and confusion among functionally close but incorrect alternatives in Hard.
Table~\ref{tab:benchmark_example} illustrates this design: Hard distractors are same-domain, same-technology, or over-generalized alternatives that remain superficially plausible but fail the required function, and serve as a targeted stress test for function-level confusion rather than an estimate of distractor prevalence in natural repositories.
The 75 core queries span 55 application domains across eight super-categories, with no single super-category exceeding 17\% (Appendix~\ref{app:eval_details}).
This targets the practically important setting of large skill registries with heavy overlap, common in community ecosystems and internal tool catalogs.

%% file: sections/body_study.tex
\section{What signals drive skill selection?}
\label{sec:body_study}

\begin{figure}[t]
\centering
\includegraphics[width=\columnwidth]{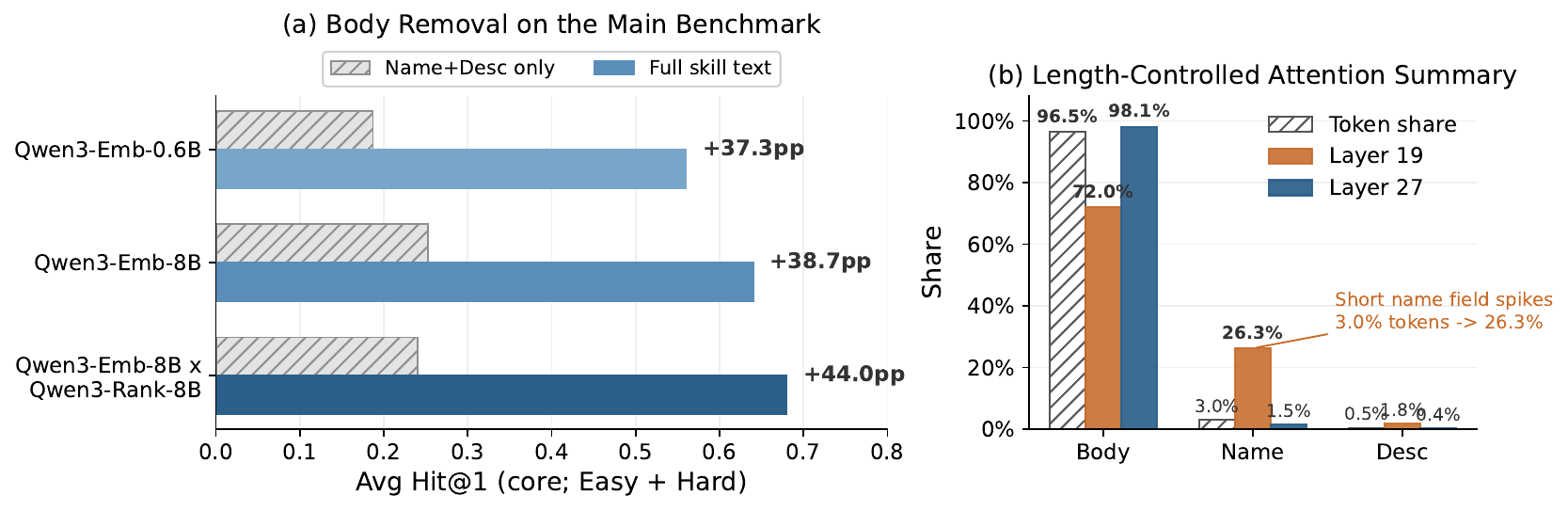}
\caption{\textbf{Body access is a critical routing signal.}
\textit{Left:} Averaged over Easy and Hard, removing body reduces Hit@1 by 37.3pp for Qwen3-Emb-0.6B, 38.7pp for Qwen3-Emb-8B, and 44.0pp for Qwen3-Emb-8B $\times$ Qwen3-Rank-8B.
\textit{Right:} A length-controlled diagnostic: the body occupies 96.5\% of skill tokens, yet the 3.0\%-token name field peaks at 26.3\% attention in layer 19 before the final layer returns to 98.1\% body attention.}
\label{fig:body_decisive}
\end{figure}

Current agent frameworks typically expose only a skill's name and description, implicitly assuming that metadata is sufficient for selection.
We test this assumption on the paper's main benchmark, reporting the Easy/Hard average used elsewhere.
We use \textbf{nd} for name+description only and \textbf{all-field} for name+description plus a body prefix capped by each model's input budget (Section~\ref{sec:method}).
Figure~\ref{fig:body_decisive} compares Qwen3-Emb-0.6B, the strongest encoder-only base model (Qwen3-Emb-8B), and the strongest base retrieve-and-rerank pipeline (Qwen3-Emb-8B $\times$ Qwen3-Rank-8B).
Appendix Table~\ref{tab:nd_full_details} reports the corresponding per-tier values.

\paragraph{Body removal collapses performance across method families.}
Qwen3-Emb-0.6B falls from 56.0\% to 18.7\%, Qwen3-Emb-8B from 64.0\% to 25.3\%, and Qwen3-Emb-8B $\times$ Qwen3-Rank-8B from 68.0\% to 24.0\%.
This collapse is therefore not tied to a single model choice: across encoder-only retrieval and reranking, removing the body removes a critical routing signal and sharply degrades top-rank performance.

\begin{table}[H]
\centering
\small
\setlength{\tabcolsep}{3.5pt}
\begin{tabular}{@{}lccc@{}}
\toprule
\textbf{System} & \textbf{Raw nd} & \textbf{Oracle nd} & \textbf{All-field} \\
\midrule
SkillRouter-1.2B pipeline & .273 & .527 & \textbf{.740} \\
Qwen3 0.6B $\times$ 0.6B & .267 & .567 & \textbf{.640} \\
Qwen3 8B $\times$ 8B & .240 & .613 & \textbf{.680} \\
\midrule
\multicolumn{4}{@{}l}{\textit{Encoder-only, matched training for each input regime}} \\
Fine-tuned, nd input & .513 & .320 & -- \\
Fine-tuned, all-field input & -- & -- & \textbf{.653} \\
\bottomrule
\end{tabular}
\caption{Stronger metadata controls on the main benchmark (Hit@1, Easy/Hard average). Oracle nd is a query-blind, body-informed replacement; Appendix~\ref{app:oracle_nd} gives its exact construction. The fine-tuned encoders are trained separately on the same 37,979 pairs; .320 applies the GT-only Oracle replacement to the nd-trained encoder.}
\label{tab:metadata_controls}
\end{table}

\paragraph{Body-distilled metadata recovers part, but not all, of the gap.}
Table~\ref{tab:metadata_controls} tests a stronger counter-hypothesis than naturally written descriptions.
For each GT skill, GPT-5.4-mini creates a query-blind, body-grounded summary from its name, description, and body.
This oracle nd raises the three pipelines from 24.0--27.3\% to 52.7--61.3\%, confirming that the missing information is body-resident rather than a simple consequence of adding tokens.
Direct all-field routing still leads by 6.7--21.3pp in the tested pipelines; on SkillBench-Supp, the corresponding residual is 5.1--12.1pp.
Appendix~\ref{app:oracle_nd} gives the exact Oracle-ND construction and the complete supplementary result.

\paragraph{Matched metadata-only fine-tuning does not close the gap.}
We also fine-tune Qwen3-Emb-0.6B on nd-only inputs with the same 37,979 query pairs and negative recipe used by the all-field encoder, withholding body in both training and inference.
Under this matched comparison, the nd-only encoder reaches 51.3\%, whereas the all-field encoder reaches 65.3\%.
Thus, withholding body in both training and inference leaves a 14.0pp gap.
For the separate GT-only Oracle-ND diagnostic, the nd-trained encoder changes from 51.3\% to 32.0\%; this intervention is not a corpus-wide metadata rewrite and should not be read as a third matched training regime.

\paragraph{Length-controlled diagnostics support the same interpretation.}
For each query--skill pair, we average the last decision token's attention over heads, sum it within name, description, and body spans, and normalize across those fields; field token share provides the length baseline.
The name field covers only 3.0\% of skill tokens yet rises to 26.3\% attention at layer 19 before the final layer returns to 98.1\% body attention.
Final-layer body attention exceeds body token share on 69/75 queries and is uncorrelated with absolute body length ($r=0.04$).
Separately, the nd$\rightarrow$all-field gap remains 26.9, 50.0, 39.5, and 31.8pp across increasing ground-truth description-length quartiles, including descriptions longer than 35 words.
These are diagnostic rather than causal results, but they argue against a pure length explanation; the oracle-nd experiment more directly tests description information quality.
Appendices~\ref{app:attention} and~\ref{app:desc_quality} provide the full layer-wise, query-level, and description-quartile analyses.

\paragraph{Implication.}
Raw community metadata omits discriminative body information.
A router can read it directly or partially compile it into richer metadata; direct all-field routing is strongest here.
This motivates \skillrouter in the next section.

%% file: sections/method.tex
\section{SkillRouter: a compact body-aware routing recipe}
\label{sec:method}

Motivated by the body-access finding in Section~\ref{sec:body_study}, we present \skillrouter, a compact body-aware retrieve-and-rerank pipeline tailored to large, homogeneous skill pools.
Its main contribution is a setting-specific routing recipe, together with two training adaptations that materially improve performance in this regime: false-negative filtering to handle near-duplicate skills that corrupt contrastive learning, and listwise reranking to resolve fine-grained competition among topically similar candidates.
We do not introduce a new encoder or reranker architecture; rather, we show that these choices are important in this setting and that the resulting compact pipeline occupies a favorable efficiency--accuracy frontier.

\begin{figure}[t]
\vspace{-10pt}
\centering
\includegraphics[width=.95\columnwidth]{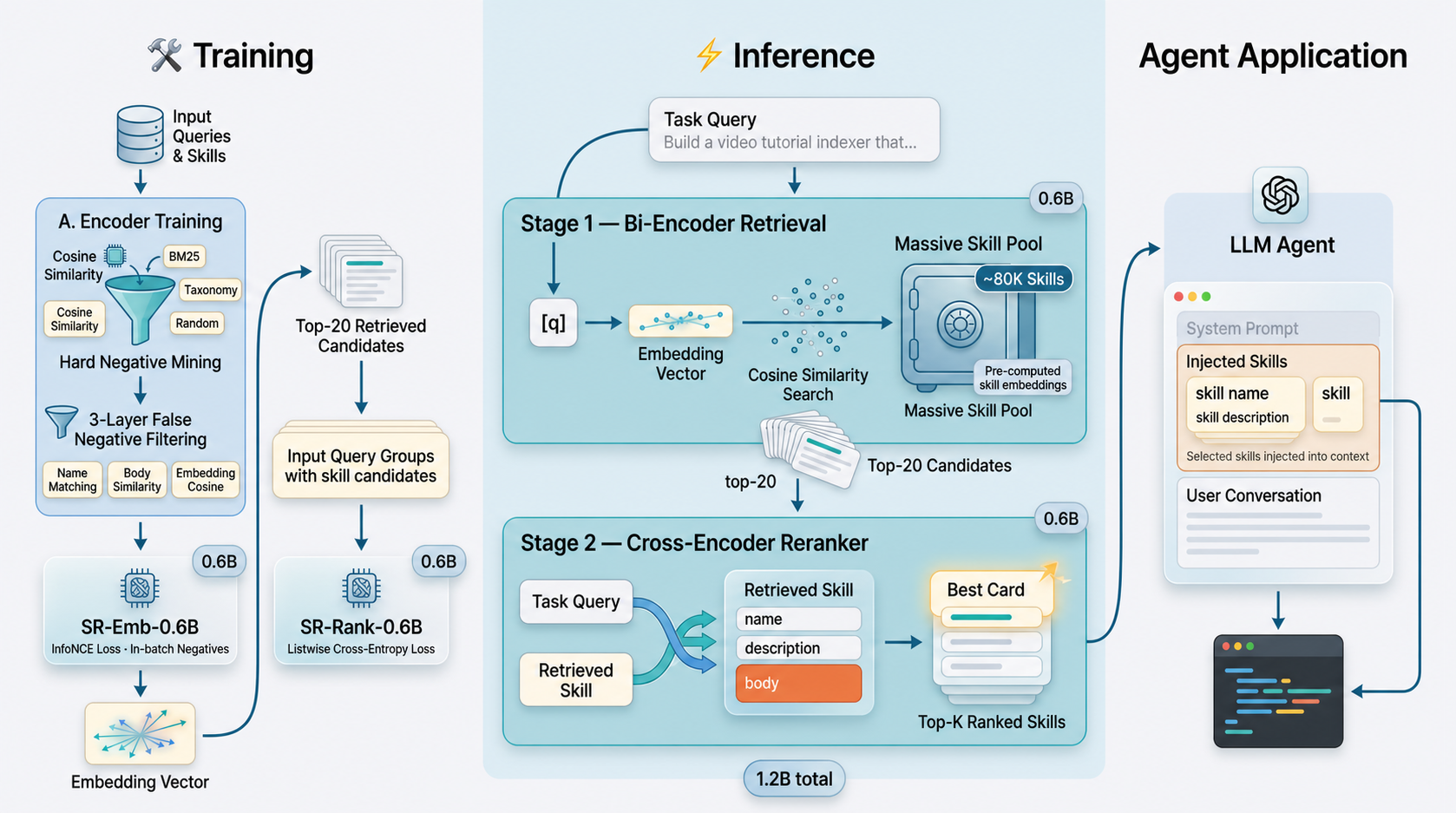}
\caption{\skillrouter pipeline. A bi-encoder retrieves top-20 candidates from the ${\sim}$80K pool; a cross-encoder reranks them. Both stages use name, description, and body content, motivated by Section~\ref{sec:body_study}.}
\label{fig:pipeline}
\end{figure}

Concretely, \skillrouter is a body-aware two-stage pipeline: a bi-encoder first retrieves a short candidate list from the ${\sim}$80K pool, and a cross-encoder then reranks those candidates using all skill fields within fixed input budgets.
Our primary configuration uses a 0.6B encoder and a 0.6B reranker, for 1.2B parameters total.
Figure~\ref{fig:pipeline} summarizes the training setup and the two-stage inference path.

\paragraph{Bi-encoder retrieval.}
\label{sec:encoder}
We fine-tune Qwen3-Emb-0.6B \citep{qwen3emb2025} on 37,979 synthetic (query, skill) pairs.
Skills are sampled from the ${\sim}$80K community pool with stratified sampling to ensure category diversity.
For each sampled skill, GPT-4o-mini generates a realistic task description from its metadata and body while being instructed not to reveal the skill name, so generated queries reflect functional need rather than lexical identity.
Benchmark-labeled skills are excluded from training supervision, ensuring the encoder learns transferable routing patterns rather than memorizing benchmark skills.
We optimize the retriever with in-batch InfoNCE over all-field skill inputs.
At inference time, the encoder embeds the full skill inventory offline and retrieves only the top-20 candidates, giving the second stage a narrow but still diverse decision set.
Appendix~\ref{app:data_construction} gives the full construction procedure and query-generation prompt.

\paragraph{Hard negative mining.}
In practice, a single user request may match dozens of superficially relevant skills---e.g., multiple ``git'' or ``docker'' management tools---while only one provides the specific capability needed.
Random negatives cannot teach the encoder to make these fine-grained distinctions.
Each query is paired with 10 negatives from four complementary sources: \emph{semantic} neighbors (4 per query) retrieved by the base encoder's embeddings, \emph{lexical} matches (3) via BM25 scoring, \emph{taxonomy} distractors (2) from the same skill category, and \emph{random} negatives (1) from a different category.
This mixture forces the encoder to distinguish semantically close alternatives, lexical confounders, and same-category distractors simultaneously, precisely where body access becomes operationally important.

\paragraph{False negative filtering.}
Because the hard negatives above are mined from a pool where the same capability is often independently implemented by different authors under different names, the mined candidate set inevitably includes skills that are functionally equivalent to the ground truth.
Treating these as negatives corrupts the contrastive signal.
We apply a three-layer filter: name deduplication, body-text overlap (trigram Jaccard $>0.6$), and embedding similarity ($>0.92$), removing approximately 10\% of mined negatives.
Section~\ref{sec:calib_ablate} shows that this filtering contributes +4.0pp Hit@1.
Appendix~\ref{app:data_construction} provides the complete mining and filtering details.

\paragraph{Cross-encoder reranking.}
\label{sec:reranker}
The retriever supplies the top-20 candidates to a fine-tuned Qwen3-Rank-0.6B \citep{qwen3emb2025}, which scores each query-skill pair using flattened all-field skill text.
Training uses 32,283 candidate lists retrieved by SR-Emb-0.6B, each containing 20 skills with binary relevance labels; the same false-negative filtering pipeline as the encoder stage is applied.
We adopt listwise cross-entropy rather than pointwise binary classification: once the retriever has narrowed the pool to 20 candidates, the remaining skills are often all topically plausible, so the reranker must compare candidates against one another rather than score each independently.
Section~\ref{sec:calib_ablate} shows that listwise training is essential, outperforming the pointwise variant by 30.7pp Hit@1.

\paragraph{Implementation details.}
Both models are trained on a single GPU.
At inference time, skills are pre-embedded offline; a live query requires one encoder forward pass, approximate nearest-neighbor search, and reranking of 20 candidates.
The encoder handles large-pool recall, while the reranker spends its all-field capacity on fine-grained distinctions among similar candidates.
The encoder caps query, description, and body text at 1,500, 300, and 2,500 characters, respectively, with a 2,048-token maximum; the reranker caps description and body at 500 and 2,000 characters with a 4,096-token maximum.
Both models train for one epoch with AdamW on a single 96GB GPU; encoder and reranker learning rates are $2{\times}10^{-5}$ and $1{\times}10^{-5}$.
A sweep over $K\in\{10,20,50\}$ selects $K{=}20$: it matches or exceeds the alternatives on both tiers while retaining sufficient candidate coverage.
Exact input templates and training settings appear in Appendix~\ref{app:templates}, and Appendix~\ref{app:topk} reports the top-$K$ ablation.

%% file: sections/experiments.tex
\section{Experiments}
\label{sec:eval}

\subsection{Setup}
\label{sec:setup}

Our primary evaluation uses the benchmark in Section~\ref{sec:problem}: 75 core queries over an ${\sim}$80K skill pool, evaluated on both Easy and Hard tiers and averaged unless otherwise noted. To assess generalization beyond this core benchmark, we report results on a supplementary benchmark in Section~\ref{sec:supp_bench}.
Our primary metric is Hit@1, with MRR@10 as a secondary ranking metric.
For multi-skill queries, we additionally report Recall@$K$ and FC@10 as coverage metrics: R@10 serves as the main shortlist-coverage metric, while FC@10 provides a stricter full-coverage view.
All models in the main results use \textit{all-field} skill inputs; metadata controls are summarized in Section~\ref{sec:body_study}.
Unless otherwise stated, rerankers operate on the encoder's top-20 candidate list.

\paragraph{Input formats and baselines.}
Each skill contains a \textit{name}, \textit{description}, and \textit{body}; we use \textbf{all-field} (all three fields, capped as specified in Section~\ref{sec:method}) and \textbf{nd} (name+description only).
All tuned models in the main result tables use all-field inputs.

\paragraph{Encoder baselines.}
We compare four encoder families:
\begin{itemize}[nosep,leftmargin=*]
\item \textbf{Sparse retrieval:} BM25 \citep{robertson2009bm25} over all-field skill inputs.
\item \textbf{Traditional open bi-encoders:} E5-Large-v2 \citep{wang2024text}, GTE-Large-v1.5 \citep{li2024gte}, and BGE-Large-v1.5 \citep{xiao2024bge}.
\item \textbf{Decoder-based encoders:} Qwen3-Emb-0.6B, Qwen3-Emb-8B \citep{qwen3emb2025}, and NV-Embed-v2 \citep{lee2024nv}.
\item \textbf{Proprietary APIs:} OpenAI \texttt{text-embedding-3-large} \citep{openai2024text} and Gemini \texttt{gemini-embedding-001} \citep{google2025gemini}.
\end{itemize}
Table~\ref{tab:encoder_results_main} reports representative models from each family; the full retrieval grid appears in Appendix~\ref{app:full_results}.

\paragraph{Reranker baselines and our systems.}
For reranking we evaluate Qwen3 base rerankers \citep{qwen3emb2025} and listwise LLM-as-judge baselines, all operating on the encoder's top-20 candidate list.
Our own systems include SR-Emb-0.6B / SR-Rank-0.6B as the primary compact pipeline, plus 8B-component scaling variants to test recipe transfer.
The benchmark stresses both stages through scale, overlap, and lexical mismatch: encoders must retrieve through category overlap and many plausible alternatives, while rerankers must sort highly similar candidates within the top-20 window.

\subsection{Main results}
\label{sec:main_results}

\begin{table*}[t]
\vspace{-10pt}
\centering
\small
\begin{tabular}{@{}llccccc@{}}
\toprule
\textbf{Model} & \textbf{Params} & \textbf{E-Hit@1} & \textbf{H-Hit@1} & \textbf{A-Hit@1} & \textbf{A-MRR@10} & \textbf{A-R@20} \\
\midrule
BM25 & -- & .347 & .280 & .314 & .365 & .365 \\
BGE-Large-v1.5 & 335M & .613 & .587 & .600 & .653 & .668 \\
gemini-embedding-001 & --  & .613 & .560 & .587 & .650 & .687 \\
text-embedding-3-large & -- & .640 & .600 & .620 & .658 & .664 \\
Qwen3-Emb-0.6B & 0.6B & .587 & .533 & .560 & .638 & .637 \\
Qwen3-Emb-8B & 8B & \underline{.653} & \underline{.627} & \underline{.640} & \underline{.698} & \underline{.726} \\
\rowcolor{blue!7}
SR-Emb-0.6B & 0.6B & \textbf{.667} & \textbf{.640} & \textbf{.653} & \textbf{.723} & \textbf{.754} \\
\gc{SR-Emb-8B} & \gc{8B} & \gc{.693} & \gc{.667} & \gc{.680} & \gc{.731} & \gc{.777} \\
\bottomrule
\end{tabular}
\caption{Encoder-only retrieval results on the 80K skill-routing benchmark. The tuned 0.6B encoder (highlighted) outperforms the 13$\times$ larger base encoder, showing that task-specific training compensates for scale in this setting. E/H/A denote Easy/Hard/Average. R@20 reflects candidate coverage for downstream reranking.}
\label{tab:encoder_results_main}
\end{table*}

\paragraph{Fine-tuning is more valuable than scale alone.}
Table~\ref{tab:encoder_results_main} shows that, among encoder-only systems, SR-Emb-0.6B reaches 65.3\% average Hit@1, improving by +9.3pp over the same-size Qwen3-Emb-0.6B base model and still edging past Qwen3-Emb-8B at 64.0\% despite a 13$\times$ parameter gap.
This indicates that, in this setting, skill-routing data and task-specific negatives can compensate for a 13$\times$ parameter gap.

\paragraph{The retriever also gives the reranker useful headroom.}
SR-Emb-0.6B reaches 75.4\% average R@20, exceeding Qwen3-Emb-8B at 72.6\%.
This matters because reranking can only help when the correct skill enters the candidate set.
The encoder improvements are therefore not just better top-1 ranking, but also better coverage for the second stage.

\begin{table*}[t]
\centering
\small
\begin{tabular}{@{}llccccc@{}}
\toprule
\textbf{Encoder} & \textbf{Reranker} & \textbf{E-Hit@1} & \textbf{H-Hit@1} & \textbf{A-Hit@1} & \textbf{A-MRR@10} & \textbf{A-R@10} \\
\midrule
Qwen3-Emb-0.6B & Qwen3-Rank-0.6B & .653 & .627 & .640 & .684 & .604 \\
Qwen3-Emb-8B & Qwen3-Rank-0.6B & .613 & .547 & .580 & .672 & .694 \\
Qwen3-Emb-8B & Qwen3-Rank-8B & .680 & .680 & .680 & .745 & .692 \\
SR-Emb-0.6B & Qwen3-Rank-0.6B & \underline{.720} & .693 & .707 & .769 & \underline{.724} \\
SR-Emb-0.6B & Qwen3-Rank-8B & \underline{.720} & \underline{.707} & \underline{.714} & \underline{.776} & \textbf{.727} \\
\rowcolor{blue!7}
SR-Emb-0.6B & SR-Rank-0.6B & \textbf{.760} & \textbf{.720} & \textbf{.740} & \textbf{.791} & .704 \\
\gc{SR-Emb-8B} & \gc{SR-Rank-8B} & \gc{.787} & \gc{.733} & \gc{.760} & \gc{.808} & \gc{.719} \\
\bottomrule
\end{tabular}
\caption{End-to-end retrieve-and-rerank results (top-20 candidates). The compact 1.2B tuned pipeline (highlighted) reaches the highest Hit@1 among non-scaling configurations, exceeding the 16B base pipeline at 13$\times$ fewer parameters. E/H/A denote Easy/Hard/Average.}
\label{tab:pipeline_results_main}
\end{table*}

\paragraph{The compact pipeline matches or exceeds the strongest base system at 13$\times$ fewer parameters.}
Table~\ref{tab:pipeline_results_main} shows that our primary 1.2B pipeline, SR-Emb-0.6B $\times$ SR-Rank-0.6B, reaches 74.0\% average Hit@1, compared with 68.0\% for the 16B strongest base pipeline (Qwen3-Emb-8B $\times$ Qwen3-Rank-8B).
It also improves by +10.0pp over the same-size 1.2B base configuration and by +8.7pp over encoder-only retrieval with the same tuned encoder.
The gain remains positive on both Easy (+8.0pp) and Hard (+4.0pp).
Combined with the serving results in Section~\ref{sec:serving_efficiency}---5.8$\times$ lower latency and 15.8\% less GPU memory---the compact pipeline occupies a favorable position on the efficiency--accuracy frontier.
Section~\ref{sec:supp_bench} further validates that the same directional advantage persists on an independently constructed supplementary benchmark.

\paragraph{Base rerankers help, but tuned reranking helps more.}
The tuned 1.2B pipeline reaches 74.0\% compared with 71.4\% for SR-Emb-0.6B $\times$ Qwen3-Rank-8B and 68.0\% for the 16B base pipeline, showing that task-specific adaptation in both stages contributes to the overall gain.
LLM-as-judge baselines are not competitive: the strongest judge (GPT-4o-mini \citep{openai2024gpt4omini}) reaches only 67.3\% Hit@1 on the same candidate lists, with GPT-5.4-mini \citep{openai2026gpt54mini} at 66.0\%; both judges provide only a top-1 choice rather than a scored full reranking.
The same training recipe also scales to 8B components: the SR-Emb-8B $\times$ SR-Rank-8B pipeline (16B total) yields 76.0\% Hit@1, though the 1.2B system already captures most of the gain. Appendix~\ref{app:full_results} reports the extended reranking table, and Appendix~\ref{app:diagnostics} gives the query-level decomposition.

\subsection{Metric calibration and key ablations}
\label{sec:calib_ablate}

\begin{table}[t]
\vspace{-10pt}
\centering
\small
\begin{tabular}{@{}lcccccc@{}}
\toprule
\textbf{Pipeline} & \multicolumn{3}{c}{\textbf{Single}} & \multicolumn{3}{c}{\textbf{Multi}} \\
\cmidrule(lr){2-4} \cmidrule(lr){5-7}
& \textbf{Hit@1} & \textbf{R@10} & \textbf{FC@10} & \textbf{Hit@1} & \textbf{R@10} & \textbf{FC@10} \\
\midrule
Qwen3-Emb-0.6B $\times$ Qwen3-Rank-0.6B & .625 & .708 & .708 & .647 & .556 & .324 \\
Qwen3-Emb-8B $\times$ Qwen3-Rank-8B & .667 & .812 & .812 & .686 & \textbf{.636} & \textbf{.382} \\
\rowcolor{blue!7}
SR-Emb-0.6B $\times$ SR-Rank-0.6B & \textbf{.729} & \textbf{.875} & \textbf{.875} & \textbf{.745} & .624 & .353 \\
\bottomrule
\end{tabular}
\caption{Single- vs.\ multi-skill calibration for two base pipelines and our primary 1.2B pipeline. Hit@1 reflects top-1 routing, R@10 reflects shortlist coverage, and FC@10 reflects strict full coverage for multi-skill queries.}
\label{tab:calibration}
\end{table}

\paragraph{Hit@1 gains should be read as top-1 routing gains.}
Table~\ref{tab:calibration} shows that the primary pipeline improves Hit@1 on both single- and multi-skill queries.
The strongest base pipeline remains better on strict multi-skill FC@10 (.382 vs.\ .353), so our main claim is strongest top-1 routing rather than uniformly better exhaustive set recovery.

\begin{table}[t]
\centering
\begingroup
\small
\begin{tabular}{@{}llccc@{}}
\toprule
\textbf{Component} & \textbf{Variant} & \textbf{Hit@1} & \textbf{MRR@10} & \textbf{R@10} \\
\midrule
\multicolumn{5}{@{}l}{\textit{Encoder training}} \\
\addlinespace[2pt]
SR-Emb-0.6B & Clean negatives & \textbf{.653} & \textbf{.723} & \textbf{.688} \\
SR-Emb-0.6B & Raw negatives & .613 & .692 & .672 \\
\midrule
\multicolumn{5}{@{}l}{\textit{Reranker training}} \\
\addlinespace[2pt]
SR-Emb-0.6B & Encoder-only (no reranking) & .653 & .723 & .688 \\
SR-Emb-0.6B $\times$ Qwen3-Rank-0.6B & Base reranker & .707 & .769 & \textbf{.724} \\
SR-Emb-0.6B $\times$ SR-Rank-0.6B (PW) & Pointwise BCE fine-tuning & .433 & .578 & .573 \\
\rowcolor{blue!7}
SR-Emb-0.6B $\times$ SR-Rank-0.6B (LW) & Listwise CE fine-tuning & \textbf{.740} & \textbf{.791} & .704 \\
\bottomrule
\end{tabular}
\endgroup
\caption{Key ablations. False-negative filtering contributes +4.0pp encoder Hit@1; listwise reranking is essential, outperforming the pointwise variant by +30.7pp. Top: encoder variants. Bottom: reranker variants using SR-Emb-0.6B as the retriever.}
\label{tab:ablations}
\end{table}

\paragraph{Two training choices are essential.}
Table~\ref{tab:ablations} shows that false-negative filtering contributes +4.0pp Hit@1 and +3.1pp MRR@10 to the encoder, and listwise training is decisive for reranking: the pointwise variant collapses to 43.3\% Hit@1 while the listwise model reaches 74.0\%.
Table~\ref{tab:calibration} complements these ablations with single- and multi-skill coverage.
Appendix Table~\ref{tab:loss_ablation} provides the full multi-metric loss comparison.

\subsection{Supplementary benchmark validation}
\label{sec:supp_bench}

To test whether the observed gains generalize beyond the 75-query core benchmark, we construct \textbf{SkillBench-Supp}, an independently built supplementary benchmark with 100 GT skills from three sources and 256 single-label LLM-generated evaluation queries over a 77K-skill pool.
Query generation uses Claude Sonnet rather than the GPT-4o-mini training generator, with different prompts and source mix; the 30 pool-selected GT skills are explicitly held out from training, and functionally equivalent pool entries are filtered before evaluation.
Using the same checkpoints without re-tuning, the compact 1.2B \skillrouter pipeline edges out the 16B Qwen3 base pipeline on Hit@1 (.641 vs.\ .637), while the 8B recipe reaches .730.
Appendix~\ref{app:supp_bench} documents benchmark construction, encoder-only results, and the Descriptive/Indirect breakdown; Appendix Table~\ref{tab:supp_oracle_nd} gives the complete Oracle-ND comparison.

\begin{table*}[t]
\centering
\small
\setlength{\tabcolsep}{4.5pt}
\begin{tabular}{@{}llcccccc@{}}
\toprule
\textbf{Encoder} & \textbf{Reranker} & \textbf{Hit@1} & \textbf{Hit@3} & \textbf{Hit@5} & \textbf{Hit@10} & \textbf{Hit@20} & \textbf{MRR@10} \\
\midrule
Qwen3-Emb-0.6B & Qwen3-Rank-0.6B & .516 & .664 & .707 & .781 & .812 & .603 \\
Qwen3-Emb-8B & Qwen3-Rank-8B & \underline{.637} & \textbf{.762} & \textbf{.801} & \textbf{.852} & \textbf{.863} & \textbf{.707} \\
SR-Emb-0.6B & Qwen3-Rank-0.6B & .574 & .707 & .762 & .805 & .828 & .652 \\
\rowcolor{blue!7}
SR-Emb-0.6B & SR-Rank-0.6B & \textbf{.641} & \underline{.746} & \underline{.781} & \underline{.809} & \underline{.840} & \underline{.699} \\
\gc{SR-Emb-8B} & \gc{Qwen3-Rank-8B} & \gc{.699} & \gc{.816} & \gc{.855} & \gc{.895} & \gc{.914} & \gc{.764} \\
\gc{SR-Emb-8B} & \gc{SR-Rank-8B} & \gc{.730} & \gc{.828} & \gc{.859} & \gc{.895} & \gc{.910} & \gc{.784} \\
\bottomrule
\end{tabular}
\caption{SkillBench-Supp retrieve-and-rerank results on 256 single-label queries over a 77K pool. The compact 1.2B tuned pipeline (highlighted) reaches the highest Hit@1 among non-scaling configurations, while gray rows show 8B-component scaling variants. Bold and underlined values denote the best and second-best non-scaling results. Checkpoints are transferred without re-tuning.}
\label{tab:supp_main}
\end{table*}

\subsection{Downstream end-to-end agent evaluation}
\label{sec:downstream_eval}

\begin{table*}[t]
\vspace{-10pt}
\centering
\small
\begin{tabular}{@{}llcccc@{}}
\toprule
\textbf{Skill Condition} & \textbf{Router / Source} & \textbf{Top-$K$} & \makecell{\textbf{Single}\\\textbf{Success}} & \makecell{\textbf{Multi}\\\textbf{Success}} & \makecell{\textbf{Overall}\\\textbf{Success}} \\
\midrule
No skills & None & -- & 12.50\% & 16.01\% & 14.89\% \\
Gold skills & Oracle ground-truth & GT & 30.90\% & 33.50\% & 32.67\% \\
Retrieved skills & Qwen3-Emb-8B $\times$ Qwen3-Rank-8B & 1 & 26.74\% & 25.33\% & 25.78\% \\
\rowcolor{blue!7}
Retrieved skills & SR-Emb-0.6B $\times$ SR-Rank-0.6B & 1 & \textbf{29.86\%} & \textbf{26.47\%} & \textbf{27.56\%} \\
Retrieved skills & Qwen3-Emb-8B $\times$ Qwen3-Rank-8B & 10 & 20.49\% & 27.78\% & 25.45\% \\
\rowcolor{blue!7}
Retrieved skills & SR-Emb-0.6B $\times$ SR-Rank-0.6B & 10 & \textbf{26.04\%} & \textbf{28.60\%} & \textbf{27.78\%} \\
\bottomrule
\end{tabular}
\caption{End-to-end agent evaluation on the 75-task core set using skills from the natural pool (without Hard-tier distractors). Results average over 3 trials $\times$ 4 coding agents. Gold skills are oracle upper bounds.}
\label{tab:downstream_main}
\end{table*}

\paragraph{Routing gains transfer to direct agent execution.}
Four coding agents---Kimi-K2.5 \citep{kimi2026k25}, glm-5 \citep{zai2026glm5}, Claude Sonnet 4.6, and Claude Opus 4.6 \citep{anthropic2026models}---run inside the Claude Code harness \citep{anthropic2025claude} with a 1200\,s timeout; the harness injects each retrieved skill's name and description into the agent context, and task setup and success criteria follow SkillsBench \citep{skillsbench2026}.
As shown in Table~\ref{tab:downstream_main}, across both top-1 and top-10 settings, \skillrouter improves average task success over the strongest base router (+1.78pp and +2.33pp, respectively), recovering about 71--73\% of the no-skill$\rightarrow$gold-skill uplift compared with 59--61\% for the base router.
Both top-1 and top-10 yield similar overall success (${\sim}$27.6--27.8\%), suggesting diminishing returns from expanding the shortlist beyond a quality threshold.
The benefit is more pronounced for stronger agents: Claude Sonnet/Opus 4.6 show an average $+$3.22pp gain, while glm-5 and Kimi-K2.5 show $+$0.89pp, consistent with a ceiling on routing utility for agents that cannot fully exploit correctly routed skills.
Appendix~\ref{app:downstream_eval} reports per-agent results and representative execution cases.

\subsection{Serving efficiency}
\label{sec:serving_efficiency}

On a real-pool GPU serving benchmark over 80 timed queries (274--5109 characters), the 1.2B \skillrouter pipeline serves the online query path at 495.8\,ms median latency---5.8$\times$ faster than the 16B base pipeline while using 15.8\% less GPU memory (Appendix~\ref{app:serving}).

%% file: sections/related_work.tex
\section{Related work}
\label{sec:related}

\paragraph{Large-scale tool and skill retrieval.}
Tool-use research spans invocation, retrieval, and execution over collections ranging from fixed tool sets to large API repositories \citep{schick2024toolformer,shen2023hugginggpt,qin2023toolllm,patil2023gorilla,du2024anytool}.
ToolRet provides 7.6K retrieval tasks over 43K heterogeneous tools \citep{shi2025toolret}; unlike its benchmark of tool/API-document retrieval, we isolate field visibility for natural-language skill bodies under progressive disclosure.
AgentSkillOS organizes and orchestrates skills through a curated capability structure and composition mechanism \citep{li2026organizing}; its metadata retrieval is complementary to our question of which skill field should be indexed.

\paragraph{Toolset construction and reranking.}
CRAFT synthesizes specialized toolsets from documentation and retrieves among them \citep{yuan2024craft}, whereas ToolRerank adaptively reranks tools using hierarchy and candidate difficulty \citep{xu2024toolrerank}.
Our standard bi-encoder/cross-encoder architecture follows neural IR \citep{karpukhin2020dense,izacard2022contriever,xiong2021ance,wang2024text,li2024gte,xiao2024bge,nogueira2020passage,sun2023rankgpt}; architectural novelty is not our claim.
We instead study long, overlapping, multi-field instructions in open community registries and quantify body access, false-negative contamination, and listwise competition.
Unlike SkillsBench \citep{skillsbench2026}, which evaluates using a provided skill, we retrieve it from roughly 80K candidates.

%% file: sections/conclusion.tex
\section{Conclusion}
\label{sec:conclusion}

At realistic registry scale, raw metadata omits critical body-resident routing signal: on our ${\sim}$80K-skill benchmark, removing body text causes 37--44pp drops across representative dense and reranking baselines, while body-distilled descriptions and matched metadata-only fine-tuning recover only part of the deficit.
A compact 1.2B body-aware retrieve-and-rerank pipeline reaches 74.0\% Hit@1, competitive with the strongest 16B base pipeline at 5.8$\times$ lower latency; false-negative filtering and listwise loss are essential in homogeneous skill pools.
The gains generalize to a supplementary benchmark (\S\ref{sec:supp_bench}) and transfer to direct task execution across four coding agents.
Whether the same field-access effect transfers to typed API repositories, specialized vertical catalogs, or agentic multi-hop retrieval remains an open question.

\paragraph{Limitations.}
The 75-query core is small; LLM-generated SkillBench-Supp is a transfer check, not broader human validation. Downstream results cover four agents under one budget without significance testing; FC@10 and top-10 success are distinct metrics.

%% file: sections/appendix.tex

\section{Evaluation details}
\label{app:eval_details}

\paragraph{Single- vs.\ multi-skill queries.}
Of the 75 core queries, 24 are \textit{single-skill} queries (exactly one ground-truth skill) and 51 are \textit{multi-skill} queries (two or more ground-truth skills required to complete the task).

\paragraph{Metric computation for multi-skill queries.}
For queries with multiple ground-truth skills $\gtset_\query = \{g_1, \ldots, g_m\}$, we define Hit@1 as the indicator of whether \textit{any} ground-truth skill appears at rank 1.
MRR@10 uses the highest-ranked ground-truth skill's reciprocal rank.
Recall@$K$ measures the fraction of ground-truth skills that appear in the top-$K$ results: $\text{R@}K = |\gtset_\query \cap \text{top-}K| / |\gtset_\query|$.
FC@10 (Full Coverage at 10) is the strictest metric: it equals 1 only when \textit{all} ground-truth skills for a query appear in the top 10. We therefore use R@10 and FC@10 to complement this any-hit criterion.

\paragraph{Core-query selection protocol.}
The underlying relevance file contains 87 SkillsBench-derived tasks.
We define the reported \textbf{75-query core benchmark} as the subset with at least one non-generic core skill (\texttt{core\_gt\_ids} non-empty), which yields 59 \texttt{clean} tasks and 16 \texttt{mixed} tasks.
The remaining 12 tasks are \texttt{generic\_only}: their labels contain only auxiliary file-type skills such as \texttt{pdf}, \texttt{docx}, \texttt{pptx}, or \texttt{xlsx}, so they are excluded from tier-specific routing metrics.

\paragraph{Metadata richness in the 80K pool.}
The poor nd results are not explained by missing descriptions alone.
In the Easy pool, descriptions are almost always present (only 0.12\% empty), but they are much shorter than full skill bodies: the median description length is 21 words, versus 704 words for the body.
Table~\ref{tab:metadata_audit} summarizes the resulting field-length distribution.

\begin{table}[t]
\centering
\begin{tabular}{@{}lc@{}}
\toprule
\textbf{Statistic} & \textbf{Value} \\
\midrule
Descriptions empty & 0.12\% \\
Descriptions $<$ 10 words & 18.66\% \\
Descriptions $<$ 25 words & 59.22\% \\
Median description length & 21 words \\
Median body length & 704 words \\
P90 body length & 1,991 words \\
\bottomrule
\end{tabular}
\caption{Metadata audit for the 78,361-skill Easy pool. Descriptions are usually present, but they are far more compressed than bodies.}
\label{tab:metadata_audit}
\end{table}

\paragraph{Query diversity.}
Table~\ref{tab:query_diversity} summarizes the diversity profile of the 75 core queries.
The queries span 55 application domains, which we group into eight super-categories for readability; the distribution is relatively balanced, with no single super-category exceeding 17\% of queries.
Within these domains, 244 unique topic tags cover areas from seismology and quantum simulation to BGP routing and video dubbing.
The SkillsBench source-task difficulty distribution is intentionally skewed toward non-trivial tasks (medium 60\%, hard 35\%, easy 5\%); this annotation is distinct from the Easy/Hard candidate-pool tiers used in our routing evaluation.
Among the 51 multi-skill queries, prior analysis identifies three structural types: complementary/pipeline tasks (43\%) where skills form sequential stages, substitute/overlap tasks (25\%) where multiple skills serve similar functions, and mixed tasks (32\%) combining both patterns.
This structural variation exercises different aspects of routing quality: pipeline queries stress recall (all stages must be found), while substitute queries stress precision (the best alternative must be ranked first).

\begin{table}[h]
\centering
\begin{tabular}{@{}ll@{}}
\toprule
\textbf{Dimension} & \textbf{Value} \\
\midrule
Application domains & 55 (8 super-categories) \\
Unique topic tags & 244 \\
SkillsBench task difficulty & easy 4 / medium 45 / hard 26 \\
Single / multi-skill & 24 / 51 \\
GT skills per query & 1--7 (mean 2.75) \\
Multi-skill types & pipeline 43\% / substitute 25\% / mixed 32\% \\
Instruction length & 36--586 words (median 169) \\
\bottomrule
\end{tabular}
\caption{Diversity profile of the 75 core benchmark queries.}
\label{tab:query_diversity}
\end{table}

\paragraph{Detailed nd/all-field results for the body-access study.}
Table~\ref{tab:nd_full_details} reports the per-tier base-model comparisons underlying the body-access study in Section~\ref{sec:body_study}.
All numbers use the same benchmark pool and the same 75 core queries as the main text.
Easy and Hard are the two evaluation tiers reported in the main paper.

\begin{table*}[t]
\centering
\begin{tabular}{@{}lllccc@{}}
\toprule
\textbf{Stage} & \textbf{Model} & \textbf{Input} & \textbf{Easy Hit@1} & \textbf{Hard Hit@1} & \textbf{Avg Hit@1} \\
\midrule
Encoder & Qwen3-Emb-0.6B & nd & .227 & .147 & .187 \\
Encoder & Qwen3-Emb-0.6B & all-field & .587 & .533 & .560 \\
Encoder & Qwen3-Emb-8B & nd & .307 & .200 & .253 \\
Encoder & Qwen3-Emb-8B & all-field & .653 & .627 & .640 \\
\midrule
Pipeline & Qwen3 0.6B $\times$ 0.6B & nd & .360 & .173 & .267 \\
Pipeline & Qwen3 0.6B $\times$ 0.6B & all-field & .653 & .627 & .640 \\
Pipeline & Qwen3 8B $\times$ 8B & nd & .293 & .187 & .240 \\
Pipeline & Qwen3 8B $\times$ 8B & all-field & .680 & .680 & .680 \\
\bottomrule
\end{tabular}
\caption{Detailed nd/all-field Hit@1 on the same benchmark pool and 75 core queries used in the main text. The rows provide the per-tier values underlying the representative comparisons in Figure~\ref{fig:body_decisive}.}
\label{tab:nd_full_details}
\end{table*}


\section{Benchmark data}
\label{app:benchmark_data}

\paragraph{Hard tier distractor generation.}
The Hard tier augments the Easy pool with 780 LLM-generated distractor skills.
For each ground-truth skill, we prompt GPT-4o-mini to generate 3--5 plausible-but-incorrect skills using three distractor strategies: \textit{same-domain-different-problem} (same technical domain but solves a different task), \textit{same-tech-different-use} (same technology stack but different application), and \textit{over-generalized} (broader version that lacks the specific capability needed).
We use these distractors as a targeted robustness stress test for function-level confusion, not as an estimate of their exact prevalence in natural repositories.
Table~\ref{tab:distractor_prompt} shows the generation prompt.

\paragraph{Pool deduplication and canonicalization.}
The 80K pool is deduplicated by skill ID only.
Exact duplicate IDs are removed as data-cleaning artifacts, but same-name overlaps with ground-truth skills are intentionally retained to preserve realistic near-duplicate confusion rather than construct an artificially conflict-free catalog.

\begin{table}[h]
\centering
\begin{tabular}{@{}p{\columnwidth}@{}}
\toprule
\textbf{System:} You are a skill document writer for a coding agent platform. You produce SKILL.md-style documents that are plausible but address a DIFFERENT problem than the reference skill. Each distractor must look like a real, useful skill document but must NOT solve the same task as the reference. \\
\midrule
\textbf{User:} I have a ground-truth skill used for the task(s): \texttt{<task\_ids>} \\[4pt]
\textbf{Reference skill} (name: \texttt{<skill\_name>}, category: \texttt{<category>}): \\
\texttt{<body\_truncated>} \\[4pt]
Generate \texttt{<num\_distractors>} HARD distractor skills. Each distractor must be a complete SKILL.md document that looks relevant to someone searching for this skill, but actually solves a different problem. \\[4pt]
Use these distractor strategies (one per distractor): \\
\textit{same-domain-diff-problem, same-tech-diff-use, over-generalized} \\[4pt]
For EACH distractor, output a JSON object with fields: \\
\texttt{distractor\_type}, \texttt{name}, \texttt{description}, \texttt{body} (400--1200 words). \\
\bottomrule
\end{tabular}
\caption{Distractor skill generation prompt for the Hard evaluation tier.}
\label{tab:distractor_prompt}
\end{table}


\section{Detailed attention analysis}
\label{app:attention}

\begin{figure}[H]
\centering
\includegraphics[width=\columnwidth]{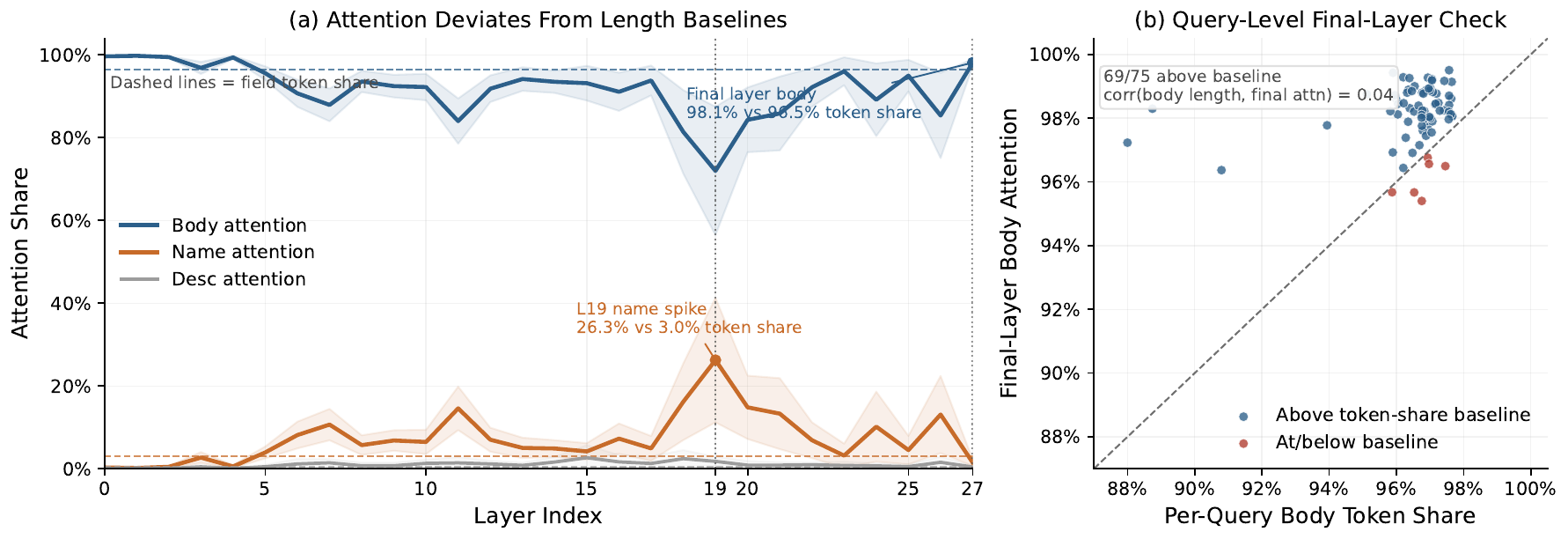}
\caption{Length-controlled attention visualization for SR-Rank-0.6B on 75 query-skill pairs. \textbf{Left:} per-layer mean attention trajectories compared against each field's token-share baseline; shaded bands denote $\pm 1$ standard deviation across the 75 query-skill pairs. The short name field spikes far above its 3.0\% token-share baseline in the middle layers, while the final layer returns to body. \textbf{Right:} query-level final-layer body attention compared against each query's body-token baseline. Most points lie above the diagonal, meaning the final layer attends to body \emph{more} than a pure length-based baseline would predict.}
\label{fig:attention_length_control}
\end{figure}

Table~\ref{tab:attention_layers} summarizes the raw per-layer attention distribution for SR-Rank-0.6B.
Because raw attention mass is length-confounded, Table~\ref{tab:attention_length_control} compares the same traces against field token-share baselines.
Figure~\ref{fig:attention_length_control} then gives the two most useful visual views of the same evidence: a layer-wise trajectory against token-share baselines and a query-level final-layer check.

\paragraph{Exact computation.}
For each analyzed query--skill pair $i$, we tokenize the flattened reranker input and identify the token spans $T_{i,f}$ of the three skill fields $f \in \{\mathrm{name}, \mathrm{desc}, \mathrm{body}\}$ in the model input.
Let $p_i$ denote the last input position, whose hidden state is used to predict the next token and thus the yes/no decision. At each layer $\ell$, we average attention from $p_i$ over heads and sum the resulting mass over the tokens in each field.
We then normalize within the three skill fields:
\[
a_{i,\ell,f}
=
\frac{\sum_{t \in T_{i,f}} \bar{A}_{i,\ell}(p_i,t)}
{\sum_{f'} \sum_{t \in T_{i,f'}} \bar{A}_{i,\ell}(p_i,t)}.
\]
The corresponding length baseline is the field's token share,
\[
b_{i,f}
=
\frac{|T_{i,f}|}{\sum_{f'} |T_{i,f'}|}.
\]
If attention were explained mainly by field length, we would expect $a_{i,\ell,f}$ to remain close to $b_{i,f}$ across layers.
Figure~\ref{fig:attention_length_control} (left) compares layer-wise mean attention shares against mean token-share baselines, while Figure~\ref{fig:attention_length_control} (right) compares final-layer body attention $a_{i,L,\mathrm{body}}$ against the per-query baseline $b_{i,\mathrm{body}}$.

\begin{table}[h]
\centering
\begin{tabular}{@{}lccc@{}}
\toprule
\textbf{Layer (Group)} & \textbf{Name} & \textbf{Desc} & \textbf{Body} \\
\midrule
\multicolumn{4}{@{}l}{\textit{Layer groups (averaged)}} \\
\addlinespace[2pt]
Early (0--6) & 2.3\% & 0.3\% & \textbf{97.3\%} \\
Middle (7--20) & 9.6\% & 1.4\% & \textbf{89.0\%} \\
Late (21--27) & 7.5\% & 0.8\% & \textbf{91.7\%} \\
\midrule
\multicolumn{4}{@{}l}{\textit{Key individual layers}} \\
\addlinespace[2pt]
Layer 0 & 0.3\% & 0.1\% & \textbf{99.6\%} \\
Layer 11 & 14.6\% & 1.4\% & 84.0\% \\
Layer 19 & \textbf{26.3\%} & 1.8\% & 72.0\% \\
Layer 27 & 1.5\% & 0.4\% & \textbf{98.1\%} \\
\midrule
Overall & 7.3\% & 1.0\% & \textbf{91.7\%} \\
\bottomrule
\end{tabular}
\caption{Raw attention distribution across skill fields by layer group and key individual layers (SR-Rank-0.6B, 28 layers $\times$ 16 heads, 75 queries).}
\label{tab:attention_layers}
\end{table}

\begin{table}[h]
\centering
\begin{tabular}{@{}lcccc@{}}
\toprule
\textbf{Field} & \textbf{Token Share} & \textbf{Overall Attn.} & \textbf{Layer 19} & \textbf{Layer 27} \\
\midrule
Name & 3.0\% & 7.3\% & \textbf{26.3\%} & 1.5\% \\
Desc & 0.5\% & 1.0\% & 1.8\% & 0.4\% \\
Body & \textbf{96.5\%} & 91.7\% & 72.0\% & \textbf{98.1\%} \\
\bottomrule
\end{tabular}
\caption{Length-controlled attention diagnostics for the same 75 query-skill pairs. Attention share is measured from the last input position used to predict the next token (and thus the yes/no decision), averaged over heads, and normalized within the name/description/body fields. If attention were explained mainly by field length, per-layer attention would remain close to field token share.}
\label{tab:attention_length_control}
\end{table}

\paragraph{Why this is not just ``more text.''}
Figure~\ref{fig:attention_length_control} extends the compact summary in Figure~\ref{fig:body_decisive} with the complete layer-wise trajectory and query-level comparison.
Final-layer body attention exceeds the body's token share on 69 of 75 queries and is effectively uncorrelated with absolute body length ($r=0.04$).
These diagnostics argue against a trivial length effect, but they are not intended as a causal test of which tokens determine the final prediction.


\section{Description-quality stratification}
\label{app:desc_quality}

A natural concern is that the nd$\rightarrow$all-field gap reported in Section~\ref{sec:body_study} could be driven by poor GT skill descriptions: if descriptions were more detailed, metadata-only routing might suffice.
We stratify the available evaluation instances by GT-description word count into four quartiles (Q1~$\leq$19 words, Q2~20--27, Q3~28--35, Q4~$>$35).
Table~\ref{tab:desc_quality_stratified} reports nd versus all-field Hit@1 per quartile for the strongest base encoder (Qwen3-Emb-8B) on the ${\sim}$80K pool.

\begin{table}[h]
\centering
\begin{tabular}{@{}lrccc@{}}
\toprule
\textbf{Quartile} & \textbf{N} & \textbf{All-field Hit@1} & \textbf{ND Hit@1} & \textbf{Gap (pp)} \\
\midrule
Q1 ($\leq$19w) & 26 & 53.8\% & 26.9\% & +26.9 \\
Q2 (20--27w) & 40 & 80.0\% & 30.0\% & +50.0 \\
Q3 (28--35w) & 38 & 65.8\% & 26.3\% & +39.5 \\
Q4 ($>$35w) & 44 & 52.3\% & 20.5\% & +31.8 \\
\midrule
Overall & 148 & 63.5\% & 25.7\% & +37.8 \\
\bottomrule
\end{tabular}
\caption{nd versus all-field Hit@1 stratified by GT-description word count. The gap remains large in every quartile, including Q4 where descriptions are longest.}
\label{tab:desc_quality_stratified}
\end{table}

The gap does not decrease monotonically with description length: Q4 (longest descriptions, $>$35 words) still shows a +31.8pp gap, and Q2 exhibits the largest gap (+50.0pp).
This pattern argues against explaining the observed gap solely through description length within the natural metadata distribution.


\section{Oracle-ND construction and supplementary results}
\label{app:oracle_nd}

\paragraph{Exact experimental definition.}
Oracle-ND is a query-blind, body-informed control over metadata quality.
For each benchmark ground-truth skill, GPT-5.4-mini receives the skill name, current author description, and full body, with the body designated as the source of truth, and generates a concise 40--60 word functional description.
Only the ground-truth skill's description field is replaced; other pool entries remain unchanged.
We compare three conditions: \textbf{raw nd} uses the as-collected name and description, \textbf{Oracle-ND} uses the name and generated description without exposing body text to the router, and \textbf{all-field} uses the name, original description, and body.
Because the replacement is GT-only, Oracle-ND is an intentionally optimistic diagnostic rather than a deployable catalog-wide metadata baseline.

\begin{quote}
\small\ttfamily
[SYSTEM]\\
You are writing the description field of a reusable software skill so that an autonomous coding agent can decide when to invoke it while browsing a large registry. You are given the skill's name, current author description, and full body. Write the best possible functional description based only on the skill's own content.\\
Rules:\\
- Describe what the skill does, when an agent should use it, and what makes it distinct.\\
- Use the skill's body as the source of truth; surface concrete, discriminative capabilities, including key operations, inputs/outputs, and domain.\\
- Do not invent capabilities not supported by the body.\\[4pt]
[USER]\\
Skill name: \textnormal{\{name\}}\\
Current author description: \textnormal{\{real description or none\}}\\
Skill body: \textnormal{\{body\}}\\[4pt]
Write a concise functional description of about 40--60 words. Capture the core function and the typical situations in which an agent should reach for this skill.
\end{quote}

For the separately nd-fine-tuned encoder reported in Table~\ref{tab:metadata_controls}, replacing raw nd with Oracle-ND changes Hit@1 from 51.3\% to 32.0\%.
This GT-only diagnostic has no corresponding all-field value for that encoder and is not directly comparable to the pipeline rows.

\paragraph{SkillBench-Supp Oracle-ND results.}
Table~\ref{tab:supp_oracle_nd} reports the complete rebuttal result on the 256-query SkillBench-Supp benchmark.

\begin{table}[H]
\centering
\small
\begin{tabular}{@{}llcccc@{}}
\toprule
\textbf{System} & \textbf{Stage} & \textbf{Raw nd} & \textbf{Oracle-ND} & \textbf{All-field} & \textbf{All-field $-$ Oracle} \\
\midrule
\multirow{2}{*}{SkillRouter-1.2B} & Encoder & 37.5 & 41.8 & 48.8 & +7.0 \\
 & Pipeline & 47.3 & 55.9 & 64.1 & +8.2 \\
\midrule
\multirow{2}{*}{Qwen3-1.2B pipeline} & Encoder & 25.4 & 35.2 & 46.1 & +10.9 \\
 & Pipeline & 32.4 & 46.5 & 51.6 & +5.1 \\
\midrule
\multirow{2}{*}{Qwen3-16B pipeline} & Encoder & 33.2 & 39.1 & 50.4 & +11.3 \\
 & Pipeline & 43.0 & 51.6 & 63.7 & +12.1 \\
\bottomrule
\end{tabular}
\caption{SkillBench-Supp Oracle-ND results (Hit@1, \%). Oracle-ND recovers part of the raw-metadata deficit, while all-field routing retains a 5.1--12.1pp pipeline advantage and a 7.0--11.3pp encoder advantage.}
\label{tab:supp_oracle_nd}
\end{table}


\section{Training data construction}
\label{app:data_construction}

This appendix provides the full details of training data construction for both stages of the \skillrouter pipeline, complementing the summary in Section~\ref{sec:method}.

\subsection{Query generation}
\label{app:query_gen}

Section~\ref{sec:method} summarizes the 37,979 synthetic (query, skill) training pairs; here we provide the generation details.
Skills are drawn from 51 categories with stratified sampling.
The generated queries have a mean length of 160 words.
Table~\ref{tab:query_prompt} shows the prompt template used with GPT-4o-mini.

\begin{table}[h]
\centering
\begin{tabular}{@{}p{\columnwidth}@{}}
\toprule
\textbf{System:} You are an experienced user of AI assistants. You write clear, realistic task requests that describe what you need to accomplish. \\
\midrule
\textbf{User:} Given the following skill specification, write a realistic task description that someone would ask an AI assistant to help with. The task should naturally require the capabilities described in this skill. \\[4pt]
Skill name: \texttt{<name>} \\
Category: \texttt{<category>} \\
Description: \texttt{<description>} \\
Skill body: \texttt{<body\_preview>} \\[4pt]
Requirements: (1) Describe a concrete scenario with specific inputs/outputs. (2) Include enough detail that the skill would be clearly useful. (3) Do NOT mention the skill name ``\texttt{<name>}'' anywhere in the task. \\[4pt]
Output ONLY the task description. \\
\bottomrule
\end{tabular}
\caption{Query generation prompt template (simplified). The LLM receives the skill's metadata and produces a realistic task description without mentioning the skill name.}
\label{tab:query_prompt}
\end{table}

\subsection{Hard negative mining}
\label{app:neg_mining}

Effective contrastive learning requires informative negatives that are challenging but not false positives.
We employ a multi-source mining strategy that produces 10 negatives per query from four complementary sources:
\begin{itemize}[nosep,leftmargin=*]
  \item \textbf{Semantic negatives} (4 per query): We pre-compute embeddings for all skills using the base Qwen3-Emb-0.6B model, retrieve the top-50 most similar skills by cosine similarity, and sample 4 non-positive skills from this set. These are the hardest negatives---semantically close but functionally distinct.
  \item \textbf{Lexical negatives} (3 per query): BM25 scoring over skill text (name + description + body) on the same top-50 candidate set, capturing term-overlap confounders that semantic search may miss.
  \item \textbf{Taxonomy negatives} (2 per query): randomly sampled from the same category as the positive skill but with a different name, providing same-domain distractors.
  \item \textbf{Random negatives} (1 per query): uniformly sampled from a different category, serving as easy negatives for calibration.
\end{itemize}

\subsection{False negative filtering}
\label{app:fn_filtering}

As described in Section~\ref{sec:method}, the mined negatives are post-processed with a three-layer filter.
The per-layer removal counts are:
\begin{enumerate}[nosep,leftmargin=*]
\item \textbf{Name deduplication}: removing negatives that share the same name as any ground-truth skill for the query (24,879 pairs removed).
\item \textbf{Body overlap}: removing negatives whose body text has trigram Jaccard similarity $>0.6$ with a ground-truth skill's body (13,860 pairs removed).
\item \textbf{Embedding similarity}: removing negatives with cosine similarity $>0.92$ to a ground-truth skill's embedding, catching semantic duplicates missed by lexical matching (326 pairs removed).
\end{enumerate}
In total, 39,065 false negative pairs are removed (approximately 10\% of all mined negative pairs).

\subsection{Reranker training data}
\label{app:reranker_data}

For each of the 32,283 training queries, we retrieve the top-20 candidates using the trained SR-Emb-0.6B encoder.
Each candidate list contains 20 skills with binary relevance labels (positive or negative), forming one training group for listwise cross-entropy optimization.
The same three-layer false negative filtering described above is applied to the reranker training data.


\section{Model input templates and training details}
\label{app:templates}

We document the exact input formats, field truncation rules, loss functions, and selected training settings used for the reported models.

\paragraph{Bi-encoder (query side).}
Following Qwen3-Emb's instruction-prefixed encoding format:
\begin{quote}
\small\ttfamily
Instruct: Given a task description, retrieve the most relevant skill document that would help an agent complete the task\\
Query: <query\_text>
\end{quote}
Before tokenization, the raw query text is truncated to 1,500 characters.

\paragraph{Bi-encoder (skill side).}
Skills are encoded as plain concatenated text without instruction prefix:
\begin{quote}
\small\ttfamily
<name> | <description> | <body>
\end{quote}
Before tokenization, description is truncated to 300 characters and body to 2,500 characters. During encoder training, each query / positive / negative input is tokenized with a maximum length of 2,048 tokens.

\paragraph{Cross-encoder reranker (flattened all-field format).}
Following the Qwen3-Rank input convention \citep{qwen3emb2025}:
\begin{quote}
\small\ttfamily
<Instruct>: Given a task description, judge whether the skill document is relevant and useful for completing the task\\[4pt]
<Query>: <query\_text>\\[4pt]
<Document>: <name> | <description> | <body>
\end{quote}
Before prompt construction, description is truncated to 500 characters and body to 2,000 characters. Tokenized reranker inputs use a maximum length of 4,096 tokens.

\paragraph{LLM-as-judge (GPT-4o-mini / GPT-5.4-mini).}
We evaluate OpenAI GPT-4o-mini and GPT-5.4-mini \citep{openai2024gpt4omini,openai2026gpt54mini} as listwise judges. Both LLM judges operate in \textit{listwise} mode: they receive the full list of top-$K$ candidates at once and select the single most relevant skill.

\textit{System prompt:}
\begin{quote}
\small\ttfamily
You are an expert at matching tasks to reusable skill definitions. Given a task query and a numbered list of candidate skills, identify the SINGLE most relevant skill that best solves the task.\\[4pt]
Respond with ONLY the number (e.g.\ `3') of the best matching skill, nothing else.
\end{quote}
\textit{User message:} The query text followed by a numbered list of candidates, each formatted as:
\begin{quote}
\small\ttfamily
[1] Name: <name>\\
Description: <description>\\
Body: <body>\\[4pt]
[2] Name: ...
\end{quote}
The selected skill is placed at rank 1; all other candidates retain their original encoder ordering.
For LLM-judge experiments, each candidate uses the same field caps as the reranker: description is truncated to 500 characters and body to 2,000 characters before prompt construction.

\paragraph{Loss definitions.}
The reported SR-Emb-0.6B model uses in-batch InfoNCE:
\begin{equation}
\mathcal{L}_{\text{enc}} = -\frac{1}{B}\sum_{i=1}^{B}
\log
\frac{\exp(\operatorname{sim}(q_i, s_i^+) / \tau)}
{\sum_{j} \exp(\operatorname{sim}(q_i, s_j) / \tau)}
\end{equation}
where $\tau = 0.05$ and $\operatorname{sim}(\cdot,\cdot)$ is cosine similarity over normalized embeddings.

The reported SR-Rank-0.6B model uses listwise cross-entropy over the top-$K$ candidate set:
\begin{equation}
\mathcal{L}_{\text{LW}} = -\log
\frac{\exp(f(q, s^+) / \tau)}
{\sum_{j=1}^{K} \exp(f(q, s_j) / \tau)}
\end{equation}
where $f(q, s)$ is the reranker score and $\tau = 1.0$ in training. For the pointwise ablation only, we instead use binary cross-entropy:
\begin{equation}
\mathcal{L}_{\text{PW}} =
-\frac{1}{K}\sum_{j=1}^{K}
\left[
y_j \log \sigma(f_j) + (1-y_j)\log(1-\sigma(f_j))
\right].
\end{equation}

\paragraph{Training hyperparameters.}
All reported training runs use a single NVIDIA GPU with 96GB HBM3 memory (Hopper architecture). Unless otherwise noted, both reported models use AdamW with weight decay 0.01, a cosine learning-rate schedule with 5\% warmup, BF16 mixed precision, and gradient checkpointing. Table~\ref{tab:train_hparams} summarizes the selected training settings for the reported models, and both reported models are described using a 1-epoch training configuration.

\begin{table*}[t]
\centering
\small
\setlength{\tabcolsep}{4pt}
\renewcommand{\arraystretch}{1.05}
\begin{tabular}{@{}>{\raggedright\arraybackslash}p{0.16\textwidth}>{\raggedright\arraybackslash}p{0.35\textwidth}>{\raggedright\arraybackslash}p{0.35\textwidth}@{}}
\toprule
\textbf{Setting} & \textbf{SR-Emb-0.6B} & \textbf{SR-Rank-0.6B} \\
\midrule
Objective & In-batch InfoNCE ($\tau{=}0.05$) & Listwise CE ($\tau{=}1.0$) \\
Input / prompt & Query instruction prefix; skill = \texttt{name | description | body} & flattened all-field query-document prompt over top-20 candidates \\
Field caps & query 1500 chars; desc 300 chars; body 2500 chars & desc 500 chars; body 2000 chars \\
Max len & 2048 & 4096 \\
Epoch & 1 & 1 \\
Batch & 8 & 1 listwise group \\
GA & 4 & 16 \\
LR & $2{\times}10^{-5}$ & $1{\times}10^{-5}$ \\
\bottomrule
\end{tabular}
\caption{Selected training hyperparameters for the reported SkillRouter models. Character caps are applied before tokenization.}
\label{tab:train_hparams}
\end{table*}


\section{\texorpdfstring{Top-$K$}{Top-K} candidate ablation}
\label{app:topk}

\begin{figure}[h]
\centering
\includegraphics[width=\columnwidth]{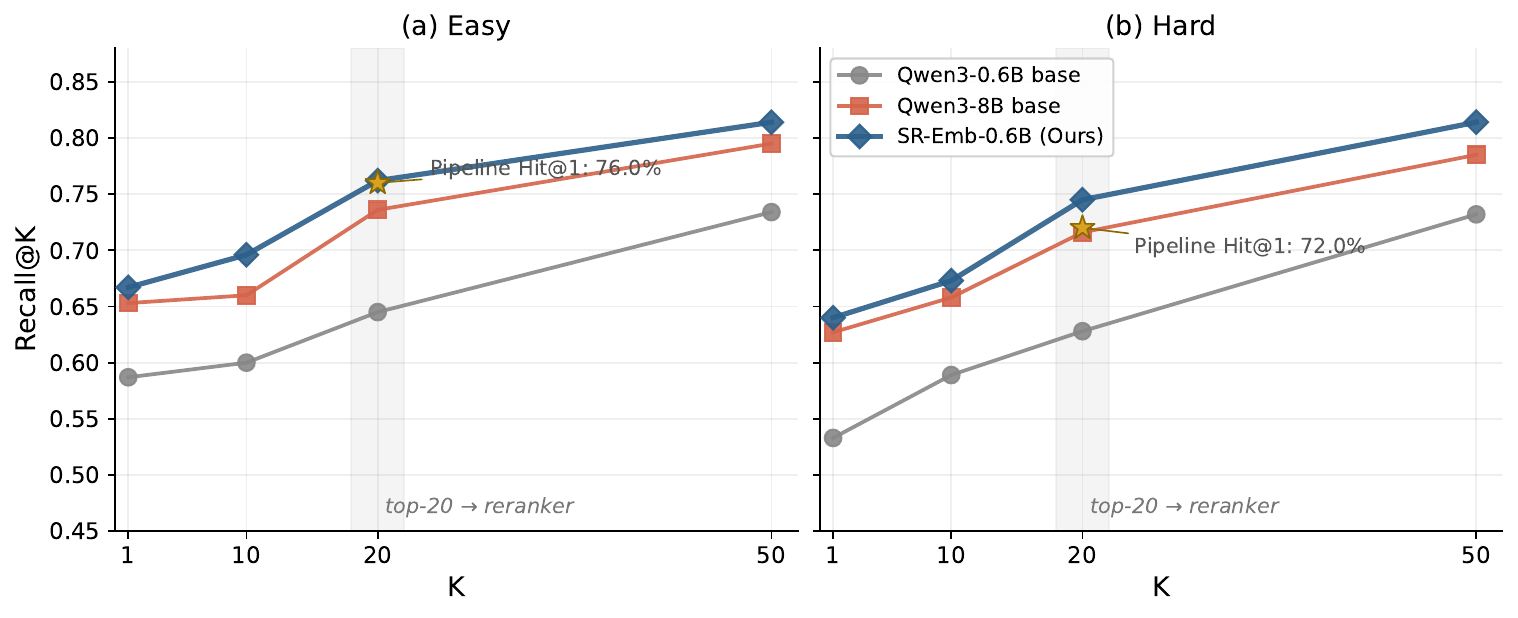}
\caption{Recall@$K$ candidate coverage for three encoder retrievers on Easy and Hard. The star marker at $K{=}20$ indicates the primary SR-Emb-0.6B $\times$ SR-Rank-0.6B pipeline's end-to-end Hit@1 at the main operating point, and is shown only as a reference against the Recall@$K$ curves.}
\label{fig:recall_ceiling_main}
\end{figure}

For the main benchmark operating point, we ablate the number of candidates ($K \in \{10, 20, 50\}$) passed from the encoder to the reranker.
Table~\ref{tab:topk} reports Hit@1 for three rerankers across both tiers.
Figure~\ref{fig:recall_ceiling_main} shows Recall@$K$ candidate coverage for three encoder retrievers, with star markers at $K{=}20$ indicating the corresponding end-to-end pipeline Hit@1.

\begin{table}[h]
\centering
\begin{adjustbox}{max width=\columnwidth}
\begin{tabular}{@{}lccccccc@{}}
\toprule
& \multicolumn{3}{c}{\textbf{Easy}} & \multicolumn{3}{c}{\textbf{Hard}} & \textbf{Avg} \\
\cmidrule(lr){2-4} \cmidrule(lr){5-7} \cmidrule(lr){8-8}
\textbf{Reranker} & @10 & @20 & @50 & @10 & @20 & @50 & @20 \\
\midrule
SR-Rank-0.6B (FT) & .747 & \textbf{.760} & .733 & .720 & \textbf{.720} & .707 & \textbf{.740} \\
Qwen3-Rank-0.6B & .720 & .720 & .667 & .693 & .693 & .640 & .707 \\
Qwen3-Rank-8B & .693 & .720 & .707 & .680 & .707 & .707 & .714 \\
\bottomrule
\end{tabular}
\end{adjustbox}
\caption{Hit@1 as a function of top-$K$ candidates for reranking on the main benchmark. Encoder = SR-Emb-0.6B. Across the reported Easy/Hard comparisons, $K{=}20$ consistently matches or exceeds the alternatives.}
\label{tab:topk}
\end{table}

Taken together, Table~\ref{tab:topk} and Figure~\ref{fig:recall_ceiling_main} motivate our choice of $K{=}20$ for the main benchmark operating point.
Figure~\ref{fig:recall_ceiling_main} shows that $K{=}20$ already captures most of the available candidate coverage for all three retrievers.
Table~\ref{tab:topk} then shows that, in this main-benchmark setting, moving to $K{=}50$ does not improve downstream Hit@1 and often hurts it, particularly for the fine-tuned SR-Rank-0.6B ($-2.0$pp average), whereas $K{=}10$ leaves less reranking headroom.


\section{Serving efficiency}
\label{app:serving}

Table~\ref{tab:serving_gpu} reports the GPU serving benchmark on the real pool.
These measurements cover the online query path only: one encoder forward pass, approximate nearest-neighbor retrieval, and top-20 reranking when applicable.
They exclude one-time model loading, offline pool embedding, and index construction costs.

\begin{table}[H]
\centering
\small
\begin{tabular}{@{}lccccc@{}}
\toprule
\textbf{System} & \textbf{p50 (ms)} & \textbf{p95 (ms)} & \textbf{QPS} & \textbf{GPU Mem (MiB)} & \textbf{GPU-sec / 1K} \\
\midrule
SR-Emb-0.6B & 19.8 & 20.8 & 50.5 & 9364 & 19.8 \\
\rowcolor{blue!7}
SR pipeline (1.2B) & \textbf{495.8} & \textbf{871.4} & \textbf{1.83} & \textbf{18976} & \textbf{547.0} \\
Qwen3-Emb-8B & 60.4 & 81.3 & 18.7 & 22539 & 53.5 \\
Qwen3 pipeline (16B) & 2900.1 & 5676.5 & 0.32 & 22539 & 6189.9 \\
\bottomrule
\end{tabular}
\caption{Real-pool GPU serving benchmark on 80 timed queries (274--5109 chars). Encoder-only rows are encoder-bound; pipeline rows are rerank-forward-bound. GPU-sec / 1K is a benchmark compute-footprint measure rather than a dollar-cost estimate.}
\label{tab:serving_gpu}
\end{table}


\section{Extended main-text tables}
\label{app:full_results}

Table~\ref{tab:app_encoder_full} provides the full encoder retrieval grid by tier and in aggregate, while Table~\ref{tab:app_rerank} gives the extended end-to-end result.
Table~\ref{tab:loss_ablation} gives the full multi-metric reranker-loss ablation.

\begin{table}[H]
\centering
\small
\setlength{\tabcolsep}{2pt}
\begin{adjustbox}{max width=\textwidth}
\begin{tabular}{@{}lllccccccccc@{}}
\toprule
\textbf{Model} & \textbf{Type} & \textbf{Params} & \textbf{E-Hit@1} & \textbf{E-R@20} & \textbf{H-Hit@1} & \textbf{H-R@20} & \textbf{A-Hit@1} & \textbf{A-MRR@10} & \textbf{A-R@10} & \textbf{A-R@20} & \textbf{A-R@50} \\
\midrule
BM25 & Sparse & -- & .347 & .376 & .280 & .354 & .314 & .365 & .321 & .365 & .433 \\
\midrule
gemini-embedding-001 & Propri. & -- & .613 & .689 & .560 & .685 & .587 & .650 & .629 & .687 & .774 \\
text-embedding-3-large & Propri. & -- & .640 & .676 & .600 & .652 & .620 & .658 & .609 & .664 & .709 \\
\midrule
E5-Large-v2 & Encoder & 335M & .507 & .594 & .493 & .594 & .500 & .565 & .553 & .594 & .622 \\
BGE-Large-v1.5 & Encoder & 335M & .613 & .677 & .587 & .658 & .600 & .653 & .608 & .668 & .743 \\
GTE-Large-v1.5 & Encoder & 434M & .573 & .706 & .520 & .686 & .547 & .631 & .630 & .696 & .750 \\
\midrule
Qwen3-Emb-0.6B & Decoder & 0.6B & .587 & .645 & .533 & .628 & .560 & .638 & .595 & .637 & .733 \\
NV-Embed-v2 & Decoder & 7B & .440 & .565 & .413 & .559 & .427 & .508 & .504 & .562 & .649 \\
Qwen3-Emb-8B & Decoder & 8B & \underline{.653} & \underline{.736} & \underline{.627} & \underline{.716} & \underline{.640} & \underline{.698} & \underline{.659} & \underline{.726} & \underline{.790} \\
\midrule
\rowcolor{blue!7}
SR-Emb-0.6B & Decoder & 0.6B & \textbf{.667} & \textbf{.762} & \textbf{.640} & \textbf{.745} & \textbf{.653} & \textbf{.723} & \textbf{.688} & \textbf{.754} & \textbf{.814} \\
\gc{SR-Emb-8B} & \gc{Decoder} & \gc{8B} & \gc{.693} & \gc{.785} & \gc{.667} & \gc{.769} & \gc{.680} & \gc{.731} & \gc{.692} & \gc{.777} & \gc{.851} \\
\bottomrule
\end{tabular}
\end{adjustbox}
\caption{Full encoder retrieval results on the 80K skill pool. All models use all-field skill input. E/H/A denote Easy/Hard/Average.}
\label{tab:app_encoder_full}
\end{table}

\begin{table}[H]
\centering
\begin{adjustbox}{max width=\textwidth}
\begin{tabular}{@{}llccccc@{}}
\toprule
\textbf{Encoder} & \textbf{Reranker} & \textbf{E-Hit@1} & \textbf{H-Hit@1} & \textbf{A-Hit@1} & \textbf{A-MRR@10} & \textbf{A-R@10} \\
\midrule
\multicolumn{7}{@{}l}{\textit{Reranker with nd input (no body)}} \\
\addlinespace[2pt]
Qwen3-Emb-8B & Qwen3-Rank-8B & .293 & .187 & .240 & .392 & .530 \\
Qwen3-Emb-0.6B & Qwen3-Rank-0.6B & .360 & .173 & .267 & .392 & .524 \\
Qwen3-Emb-8B & Qwen3-Rank-0.6B & .293 & .133 & .213 & .385 & .603 \\
Qwen3-Emb-8B & GPT-4o-mini & .213 & .173 & .193 & -- & -- \\
Qwen3-Emb-0.6B & GPT-4o-mini & .253 & .160 & .207 & -- & -- \\
Qwen3-Emb-8B & GPT-5.4-mini & .347 & .267 & .307 & -- & -- \\
Qwen3-Emb-0.6B & GPT-5.4-mini & .373 & .293 & .333 & -- & -- \\
\midrule
\multicolumn{7}{@{}l}{\textit{Reranker with all-field input: base models}} \\
\addlinespace[2pt]
Qwen3-Emb-8B & GPT-4o-mini & .667 & .627 & .647 & -- & -- \\
Qwen3-Emb-0.6B & GPT-4o-mini & .560 & .547 & .554 & -- & -- \\
Qwen3-Emb-8B & GPT-5.4-mini & .627 & .560 & .594 & -- & -- \\
Qwen3-Emb-0.6B & GPT-5.4-mini & .573 & .547 & .560 & -- & -- \\
Qwen3-Emb-0.6B & Qwen3-Rank-0.6B & .653 & .627 & .640 & .684 & .604 \\
Qwen3-Emb-8B & Qwen3-Rank-0.6B & .613 & .547 & .580 & .672 & .694 \\
Qwen3-Emb-8B & Qwen3-Rank-8B & .680 & .680 & .680 & .745 & .692 \\
\midrule
\multicolumn{7}{@{}l}{\textit{SR-Emb-0.6B + reranker (all-field input)}} \\
\addlinespace[2pt]
SR-Emb-0.6B & GPT-5.4-mini & .667 & .653 & .660 & -- & -- \\
SR-Emb-0.6B & GPT-4o-mini & .693 & .653 & .673 & -- & -- \\
SR-Emb-0.6B & Qwen3-Rank-0.6B & .720 & .693 & .707 & .769 & .724 \\
SR-Emb-0.6B & Qwen3-Rank-8B & .720 & .707 & .714 & .776 & \textbf{.727} \\
\rowcolor{blue!7}
SR-Emb-0.6B & SR-Rank-0.6B & \textbf{.760} & \textbf{.720} & \textbf{.740} & \textbf{.791} & .704 \\
\midrule
\multicolumn{7}{@{}l}{\textit{Scaling variants (8B components)}} \\
\addlinespace[2pt]
\gc{SR-Emb-0.6B} & \gc{SR-Rank-8B} & \gc{.787} & \gc{.707} & \gc{.747} & \gc{.804} & \gc{.707} \\
\gc{SR-Emb-8B} & \gc{SR-Rank-8B} & \gc{.787} & \gc{.733} & \gc{.760} & \gc{.808} & \gc{.719} \\
\bottomrule
\end{tabular}
\end{adjustbox}
\caption{Full encoder$\times$reranker results (top-20 candidates). Group headings indicate the input format, so the separate rank-input column is omitted. LLM judges only provide a top-1 choice, so only Hit@1 is reported for those rows.}
\label{tab:app_rerank}
\end{table}

\begin{table}[H]
\centering
\small
\begin{adjustbox}{max width=\textwidth}
\begin{tabular}{@{}llccccccc@{}}
\toprule
& & \multicolumn{2}{c}{\textbf{Easy}} & \multicolumn{2}{c}{\textbf{Hard}} & \multicolumn{3}{c}{\textbf{Average}} \\
\cmidrule(lr){3-4} \cmidrule(lr){5-6} \cmidrule(lr){7-9}
\textbf{Reranker} & \textbf{Loss} & Hit@1 & MRR@10 & Hit@1 & MRR@10 & Hit@1 & MRR@10 & FC@10 \\
\midrule
\textbf{SR-Rank-0.6B (LW)} & \textbf{LW} & \textbf{.760} & \textbf{.809} & \textbf{.720} & \textbf{.773} & \textbf{.740} & \textbf{.791} & .520 \\
\midrule
Qwen3-Rank-0.6B & base & .720 & .780 & .693 & .758 & .707 & .769 & \textbf{.527} \\
Qwen3-Rank-8B & base & .720 & .781 & .707 & .771 & .714 & .776 & .527 \\
\midrule
SR-Rank-0.6B (PW) & PW & .453 & .592 & .413 & .564 & .433 & .578 & .320 \\
\midrule
\multicolumn{2}{@{}l}{\textit{Encoder-only (no reranking)}} & .667 & .735 & .640 & .710 & .653 & .723 & .480 \\
\bottomrule
\end{tabular}
\end{adjustbox}
\caption{Full reranker-loss ablation. Encoder = SR-Emb-0.6B, top-20 candidates. LW = listwise cross-entropy; PW = pointwise binary cross-entropy; base = untuned.}
\label{tab:loss_ablation}
\end{table}


\section{Additional pipeline diagnostics}
\label{app:diagnostics}

\begin{table}[H]
\centering
\begin{adjustbox}{max width=\columnwidth}
\begin{tabular}{@{}lcccccc@{}}
\toprule
& \multicolumn{2}{c}{\textbf{Easy}} & \multicolumn{2}{c}{\textbf{Hard}} & \multicolumn{2}{c}{\textbf{All}} \\
\cmidrule(lr){2-3} \cmidrule(lr){4-5} \cmidrule(lr){6-7}
\textbf{Category} & $n$ & \% & $n$ & \% & $n$ & \% \\
\midrule
Both correct      & 47 & 62.7 & 45 & 60.0 & 92 & 61.3 \\
Reranker fixed    & 10 & 13.3 &  9 & 12.0 & 19 & 12.7 \\
Reranker degraded &  3 &  4.0 &  3 &  4.0 &  6 &  4.0 \\
Both missed       & 15 & 20.0 & 18 & 24.0 & 33 & 22.0 \\
\midrule
Encoder Hit@1     & 50 & 66.7 & 48 & 64.0 & 98 & 65.3 \\
Pipeline Hit@1    & 57 & 76.0 & 54 & 72.0 & 111 & 74.0 \\
\bottomrule
\end{tabular}
\end{adjustbox}
\caption{Per-query Hit@1 decomposition: SR-Emb-0.6B encoder-only vs.\ the full SR-Emb-0.6B $\times$ SR-Rank-0.6B pipeline.}
\label{tab:contribution_main}
\end{table}

\paragraph{Reranker contribution decomposition.}
Table~\ref{tab:contribution_main} shows that, across 150 Easy+Hard evaluations, the reranker fixes 19 cases (12.7\%) where the encoder misses rank 1 but still retrieves the correct skill into the top-20 window, while degrading only 6 cases (4.0\%).
The net gain is therefore +8.7pp Hit@1, from 65.3\% to 74.0\%.
The remaining 33 misses are mainly recall failures or cases that require multi-hop prerequisite inference.


\section{Case studies}
\label{app:cases}

We present five representative cases analyzing the behavior of the \skillrouter pipeline.
The main text focuses on aggregate results; here we provide five detailed success and failure cases.

\paragraph{Case 1: Reranker rescue (simpo-code-reproduction).}
\textit{Query:} Reproduce a research paper's loss function and set up the development environment. \\
\textit{GT Skill:} \texttt{nlp-research-repo-package-installment} (Python environment setup). \\
\textit{Analysis:} All encoders miss this subtle match (SR-Emb-0.6B: rank 13; base encoders: rank $>$50). Since rank 13 is within the top-20 window, the cross-encoder reranker identifies the alignment between ``setup the environment'' and the skill's dependency installation instructions, promoting GT to rank 1.

\paragraph{Case 2: Encoder advantage (workflow-automation).}
\textit{Query:} Automate a multi-step CI/CD pipeline with conditional stage execution. \\
\textit{GT Skill:} \texttt{github-actions-workflow} (GitHub Actions YAML generation). \\
\textit{Analysis:} Base encoders retrieve generic automation skills (``task-scheduler'', ``cron-manager''). SR-Emb-0.6B captures the ``CI/CD + conditional'' $\rightarrow$ ``GitHub Actions'' mapping, ranking GT at position 1 vs.\ position 8 for Qwen3-Emb-8B. The skill body explicitly describes conditional workflow syntax, which the fine-tuned model has learned to associate with CI/CD queries.

\paragraph{Case 3: Reranker rescue (data-format-conversion).}
\textit{Query:} Convert a legacy XML configuration to modern TOML format with schema validation. \\
\textit{GT Skill:} \texttt{config-format-converter} (multi-format config file conversion). \\
\textit{Analysis:} The encoder retrieves XML-focused and TOML-focused tools separately but misses the unified converter (SR-Emb-0.6B: rank 11). The reranker, through cross-attention over the body's supported format list (XML, YAML, JSON, TOML, INI), identifies the correct multi-format skill and promotes it to rank 1.

\paragraph{Case 4: Pointwise loss degradation (api-documentation).}
\textit{Query:} Generate REST API documentation from OpenAPI spec with interactive examples. \\
\textit{GT Skill:} \texttt{openapi-doc-generator} (Swagger/OpenAPI documentation tool). \\
\textit{Analysis:} SR-Emb-0.6B correctly retrieves GT at rank 1. However, SR-Rank-0.6B (PW) \textit{degrades} it to rank 18, while SR-Rank-0.6B (LW) maintains rank 1. The pointwise model assigns similar scores (${\sim}$0.52) to all API-related candidates, effectively randomizing the order. This case exemplifies why pointwise scoring fails in homogeneous candidate pools.

\paragraph{Case 5: System limitation (invoice-fraud-detection).}
\textit{Query:} Analyze invoice images to detect potential fraud patterns (duplicate amounts, suspicious vendor names). \\
\textit{GT Skill:} \texttt{pdf-table-extractor} (structured data extraction from PDF/images). \\
\textit{Analysis:} All methods fail. The connection between ``invoice fraud detection'' and ``PDF table extraction'' requires multi-hop reasoning: fraud analysis $\rightarrow$ structured data needed $\rightarrow$ invoices are PDFs $\rightarrow$ table extraction. No retrieval method in our pipeline captures this chain. The top results are fraud-detection analytics tools and image classifiers, none of which provide the prerequisite data extraction step. This case illustrates one important source of the remaining miss rate in our current evaluation: retrieval-based pipelines still struggle on tasks that require multi-hop prerequisite inference.


\section{Downstream evaluation details}
\label{app:downstream_eval}

We report the full direct execution results underlying the averaged main-text table.
All runs use the same execution harness on the 75-task core set with a 1200\,s timeout.
This appendix reports the same \textit{complementary end-to-end study using the natural pool} described in the main text, not a tier-matched continuation of the Easy/Hard retrieval benchmark.
Retrieved skills come from the natural pool, i.e., the non-synthetic benchmark pool without Hard-tier distractors.
Once the exposed skill package is fixed, the downstream execution protocol otherwise follows SkillsBench \citep{skillsbench2026}, including task setup and success validation.
The four evaluated agents are Kimi-K2.5 \citep{kimi2026k25}, glm-5 \citep{zai2026glm5}, Claude Sonnet 4.6, and Claude Opus 4.6 \citep{anthropic2026models}.
Table~\ref{tab:downstream_details} reports the per-agent success rates underlying the averaged main-text result in Table~\ref{tab:downstream_main}.

\begin{table*}[t]
\centering
\small
\begin{tabular}{@{}lllcccc@{}}
\toprule
\textbf{Model} & \textbf{Skill Condition} & \textbf{Router / Source} & \textbf{Top-$K$} & \textbf{Single} & \textbf{Multi} & \textbf{Overall} \\
\midrule
\multirow{6}{*}{Kimi-K2.5} & No skills & None & -- & 12.50 & 9.15 & 10.22 \\
 & Gold skills & Oracle ground-truth & GT & 20.83 & 25.49 & 24.00 \\
 & Retrieved skills & Qwen3-Pipeline-16B & 1 & 20.83 & 11.76 & 14.67 \\
 & Retrieved skills & SkillRouter-1.2B & 1 & 20.83 & 16.34 & 17.78 \\
 & Retrieved skills & Qwen3-Pipeline-16B & 10 & 16.67 & 21.57 & 20.00 \\
 & Retrieved skills & SkillRouter-1.2B & 10 & 19.44 & 17.65 & 18.22 \\
\midrule
\multirow{6}{*}{glm-5} & No skills & None & -- & 11.11 & 12.42 & 12.00 \\
 & Gold skills & Oracle ground-truth & GT & 37.50 & 23.53 & 28.00 \\
 & Retrieved skills & Qwen3-Pipeline-16B & 1 & 36.11 & 19.61 & 24.89 \\
 & Retrieved skills & SkillRouter-1.2B & 1 & 36.11 & 16.99 & 23.11 \\
 & Retrieved skills & Qwen3-Pipeline-16B & 10 & 19.44 & 21.57 & 20.89 \\
 & Retrieved skills & SkillRouter-1.2B & 10 & 30.56 & 22.22 & 24.89 \\
\midrule
\multirow{6}{*}{Claude Sonnet 4.6} & No skills & None & -- & 13.89 & 22.22 & 19.56 \\
 & Gold skills & Oracle ground-truth & GT & 40.28 & 43.79 & 42.67 \\
 & Retrieved skills & Qwen3-Pipeline-16B & 1 & 31.94 & 34.64 & 33.78 \\
 & Retrieved skills & SkillRouter-1.2B & 1 & 36.11 & 37.25 & 36.89 \\
 & Retrieved skills & Qwen3-Pipeline-16B & 10 & 33.33 & 38.56 & 36.89 \\
 & Retrieved skills & SkillRouter-1.2B & 10 & 36.11 & 42.48 & 40.44 \\
\midrule
\multirow{6}{*}{Claude Opus 4.6} & No skills & None & -- & 12.50 & 20.26 & 17.78 \\
 & Gold skills & Oracle ground-truth & GT & 25.00 & 41.18 & 36.00 \\
 & Retrieved skills & Qwen3-Pipeline-16B & 1 & 18.06 & 35.29 & 29.78 \\
 & Retrieved skills & SkillRouter-1.2B & 1 & 26.39 & 35.29 & 32.44 \\
 & Retrieved skills & Qwen3-Pipeline-16B & 10 & 12.50 & 29.41 & 24.00 \\
 & Retrieved skills & SkillRouter-1.2B & 10 & 18.06 & 32.03 & 27.56 \\
\bottomrule
\end{tabular}
\caption{Direct end-to-end agent success rates (\%) by model and skill condition. Each task is evaluated three times per condition. Retrieved skills use the natural pool, i.e., the non-synthetic benchmark pool without Hard-tier distractors.}
\label{tab:downstream_details}
\end{table*}

\subsection{Representative downstream cases}
\label{app:downstream_cases}

These cases illustrate how routing quality affects end-to-end agent execution.
They are mechanism illustrations drawn from the 3-trial evaluation, not statistical proofs.

\paragraph{Case A (positive): Semantic distractor avoidance (software-dependency-audit).}
\textit{Query:} Audit project dependencies for known vulnerabilities, extract CVSS scores, and produce a CSV report. \\
\textit{Gold skills:} \texttt{cvss-score-extraction}, \texttt{trivy-offline-vulnerability-scanning}, \texttt{vulnerability-csv-reporting}. \\
\textit{Retrieval:} The baseline top-1 retrieves \texttt{dependency-security} (a community-contributed skill about general dependency security), while \skillrouter top-1 retrieves \texttt{trivy-offline-vulnerability-scanning} (a gold skill providing the specific offline scanning workflow). \\
\textit{Result:} Baseline top-1 scores 0/12 across all four agents; \skillrouter top-1 scores 12/12.
Every agent transitions from complete failure to complete success when the correct skill is routed.
This is the most dramatic case in the dataset and illustrates how semantically similar but functionally distinct skills can completely block downstream execution when the wrong one is selected.

\paragraph{Case B (positive): Surface-matching trap (video-tutorial-indexer).}
\textit{Query:} Index a video tutorial by extracting and timestamping its spoken content. \\
\textit{Gold skill:} \texttt{speech-to-text}. \\
\textit{Retrieval:} Qwen3-Emb-0.6B and Qwen3-Emb-8B rank the gold skill at positions 25 and 6, respectively, whereas SR-Emb-0.6B ranks it first. In the downstream comparison, the baseline top-1 retrieves \texttt{video-explorer} (matching the surface keyword ``video''), while \skillrouter retrieves \texttt{speech-to-text} (matching the required function: transcription). \\
\textit{Result:} Baseline top-1 scores 0/12; \skillrouter top-1 scores 9/12 (glm-5 3/3, Kimi-K2.5 2/3, Opus 3/3, Sonnet 1/3).
The baseline retriever matches on the task's surface topic (video) rather than its functional requirement (audio transcription), a failure mode that the trained router avoids.

\paragraph{Case C (negative): Specialized multi-skill domains (hvac-control).}
\textit{Query:} Design and tune a model-predictive HVAC controller with safety interlocks. \\
\textit{Gold skills:} \texttt{excitation-signal-design}, \texttt{first-order-model-fitting}, \texttt{imc-tuning-rules}, \texttt{safety-interlocks}, \texttt{scipy-curve-fit}. \\
\textit{Retrieval:} The baseline top-1 retrieves \texttt{first-order-model-fitting} (a gold skill); \skillrouter top-1 retrieves \texttt{simulation-metrics} (a related but non-gold control-engineering skill).
In the top-10 setting, the baseline recovers all 5 gold skills, while \skillrouter recovers only 1. \\
\textit{Result:} Baseline top-1 scores 8/12; \skillrouter top-1 scores 4/12.
This case illustrates where the compact router falls short: in highly specialized multi-skill engineering domains requiring precise vocabulary matching across multiple sub-disciplines, the 16B baseline's deeper domain representation produces substantially better retrieval.


\section{Supplementary benchmark: SkillBench-Supp}
\label{app:supp_bench}

This appendix provides the full construction details, query-generation prompts, quality assurance measures, and complete results for the supplementary benchmark referenced in Section~\ref{sec:supp_bench}.

\subsection{GT skill sources}
\label{app:supp_skill_sources}

The 100 GT skills are drawn from three independent sources to ensure breadth:

\begin{itemize}[nosep,leftmargin=*]
\item \textbf{OpenClaw official repository} (51 skills): Built-in skills from the official OpenClaw skill repository \citep{openclaw2026repo}, covering communication, productivity, IoT, media, security, and other domains. Each skill is a standard SKILL.md document with name, description, and body fields.
\item \textbf{AgentSkillOS benchmark} (19 skills): GT skills from the AgentSkillOS evaluation set \citep{li2026organizing}, covering document processing, visualization, frontend design, and similar domains. These are included to test consistency with existing skill benchmarks.
\item \textbf{Pool-selected skills} (30 skills): Stratified-sampled from the 78K base pool across 25 application domains (gaming, travel, security, data science, etc.), selecting only uniquely named skills with body length between 1{,}000--15{,}000 characters to ensure quality and domain diversity.
\end{itemize}

\subsection{Pool construction}
\label{app:supp_pool}

The candidate pool is derived from the same ${\sim}$78K base pool used in the main benchmark.
To prevent false negatives---where a pool entry is functionally identical to a GT skill but scored as incorrect---we apply two deduplication steps:
\begin{enumerate}[nosep,leftmargin=*]
\item \textbf{Content deduplication}: Entries sharing identical or near-identical content with any GT skill are removed.
\item \textbf{Functional-overlap removal}: For each GT skill, we retrieve the top-5 most similar pool entries via BM25 (using name + description + body as the document), then apply an LLM-based equivalence judgment (Claude Sonnet) to each pair. Entries judged as functionally equivalent are removed. This process identifies and removes \textbf{114 equivalent pairs} across 49 GT skills (with \texttt{github}, \texttt{docx}, \texttt{xlsx}, and \texttt{pptx} each having 5 equivalent entries---the maximum observed).
\end{enumerate}

\noindent The final pool contains \textbf{77{,}537 skills}.

\subsection{Query generation}
\label{app:supp_query_gen}

For each GT skill, 2--3 evaluation queries are generated using Claude Sonnet (\texttt{claude-sonnet-4-6}) at two difficulty levels, with \texttt{temperature\,=\,0.8}:

\paragraph{Descriptive queries (168 total).}
These describe a concrete usage scenario without naming the skill. Two sub-styles ensure stylistic diversity:

\textit{Scenario sub-style} (82 queries, 80--250 words): Detailed task descriptions with specific context.

\begin{quote}
\small\ttfamily
Given this skill specification, write a realistic user query describing a task that requires this skill's capabilities.\\[4pt]
Skill name: \textnormal{\{name\}}\\
Description: \textnormal{\{description\}}\\
Skill content preview: \textnormal{\{body\}}\\[4pt]
Requirements:\\
1. Describe a specific task or problem the user wants to solve\\
2. Include enough context for the skill to be clearly the right one\\
3. Write naturally --- as a user would describe their need to an AI assistant\\
4. Do NOT follow a rigid structure (no mandatory file paths, no mandatory numbered lists)\\
5. Total length: 80--250 words\\
6. Do NOT mention the skill name "\textnormal{\{name\}}" or reference this specification document\\
7. Do NOT use any unique identifiers or CLI command names from the skill content\\[4pt]
Output ONLY the user query. No preamble, no explanation.
\end{quote}

\textit{Developer sub-style} (86 queries, 40--120 words): Concise, natural requests.

\begin{quote}
\small\ttfamily
Given this skill specification, write a concise user query from someone who needs this capability but doesn't know the specific skill exists.\\[4pt]
Skill name: \textnormal{\{name\}}\\
Description: \textnormal{\{description\}}\\
Skill content preview: \textnormal{\{body\}}\\[4pt]
Requirements:\\
1. Write as a user naturally asking for help --- could be casual or formal\\
2. Describe what they want to achieve, not what skill to use\\
3. Be specific enough that this skill is clearly the best match\\
4. Total length: 40--120 words\\
5. Do NOT mention the skill name "\textnormal{\{name\}}" or any CLI commands from the skill\\
6. Do NOT reference this specification document\\[4pt]
Output ONLY the user query.
\end{quote}

\paragraph{Indirect queries (88 total).}
These describe a high-level need requiring capability inference to connect to the skill (50--180 words).

\begin{quote}
\small\ttfamily
Given this skill specification, write a user query that describes a high-level need which this skill would address, without revealing the skill itself.\\[4pt]
Skill name: \textnormal{\{name\}}\\
Description: \textnormal{\{description\}}\\
Skill content preview: \textnormal{\{body\}}\\[4pt]
Requirements:\\
1. Describe a real-world problem or goal, not a skill request\\
2. The user should not know this specific skill exists --- they're describing their situation\\
3. Do NOT mention the skill name "\textnormal{\{name\}}" or any specific CLI commands from the skill\\
4. Avoid using keywords that directly appear in the skill's name or description\\
5. The connection between the query and this skill should require understanding the skill's capabilities\\
6. Total length: 50--180 words\\
7. Be specific enough that this skill is the clear best answer (not too vague)\\[4pt]
Output ONLY the user query.
\end{quote}

\paragraph{Automated quality control.}
Each generated query undergoes three checks: (1)~skill-name leakage detection, (2)~CLI command-name leakage detection, and (3)~length compliance. Non-conforming queries are automatically regenerated (up to 3 attempts). The final dataset contains 256 queries (query lengths: min\,=\,43, max\,=\,226, mean\,=\,117 words).

\subsection{Quality assurance}
\label{app:supp_quality}

Several design choices ensure that SkillBench-Supp provides a credible, unbiased evaluation:

\begin{enumerate}[nosep,leftmargin=*]
\item \textbf{Distinct generation methodology}: Evaluation queries are generated by Claude Sonnet (\texttt{claude-sonnet-4-6}), whereas the \skillrouter training queries are generated by GPT-4o-mini with a different prompt design and skill sampling strategy. This separation prevents any overlap between evaluation queries and training data in terms of model, prompt template, and source distribution.
\item \textbf{Automated leakage filtering}: Each query is checked for three types of leakage: (a)~skill-name substring match, (b)~CLI command-name match against the skill body, and (c)~length compliance with the target difficulty level. Non-conforming queries are regenerated up to 3 times before being discarded.
\item \textbf{Functional-overlap removal}: As described in Section~\ref{app:supp_pool}, 114 functionally equivalent pool entries are identified and removed via BM25 + LLM-based equivalence judgment, preventing false-negative contamination in evaluation metrics.
\item \textbf{GT-skill isolation}: The 30 pool-selected GT skills are explicitly excluded from the training GT skill set, so the supplementary benchmark does not reuse training positive skills from the base pool.
\item \textbf{Multi-source GT skills}: The 100 GT skills span three independent sources (official repository, existing benchmark, and pool selection) across 25+ application domains, reducing the risk of systematic bias toward any particular skill type or domain.
\end{enumerate}

\subsection{Full results}
\label{app:supp_full_results}

Table~\ref{tab:supp_full_combined} reports the full encoder-only results on the 77K pool with 256 single-label queries, complementing the pipeline comparison in Table~\ref{tab:supp_main}.

\begin{table}[H]
\centering
\small
\begin{adjustbox}{max width=\textwidth}
\begin{tabular}{@{}lccccccc@{}}
\toprule
\textbf{System} & \textbf{H@1} & \textbf{H@3} & \textbf{H@5} & \textbf{H@10} & \textbf{H@20} & \textbf{R@50} & \textbf{MRR@10} \\
\midrule
BM25 & .066 & .094 & .125 & .141 & .199 & .250 & .088 \\
Qwen3-Emb-0.6B & .461 & .578 & .633 & .707 & .773 & .840 & .537 \\
Qwen3-Emb-8B & .504 & .609 & .676 & .742 & .809 & .867 & .575 \\
SR-Emb-0.6B & .488 & .617 & .668 & .723 & .797 & .848 & .566 \\
\textbf{SR-Emb-8B} & \textbf{.605} & \textbf{.734} & \textbf{.762} & \textbf{.824} & \textbf{.871} & \textbf{.918} & \textbf{.675} \\
\bottomrule
\end{tabular}
\end{adjustbox}
\caption{SkillBench-Supp encoder-only results (77K pool, 256 single-label queries). The corresponding retrieve-and-rerank comparison appears in Table~\ref{tab:supp_main}.}
\label{tab:supp_full_combined}
\end{table}

\subsection{Difficulty breakdown}
\label{app:supp_breakdown}

Tables~\ref{tab:supp_breakdown_desc} and~\ref{tab:supp_breakdown_ind} report the Descriptive and Indirect difficulty breakdowns for both encoder-only and pipeline systems.
Indirect queries are consistently harder across all systems, with the largest difficulty gap on BM25 (H@1 drops from .083 to .034), confirming that Indirect queries require deeper semantic understanding.

\begin{table}[H]
\centering
\small
\begin{adjustbox}{max width=\textwidth}
\begin{tabular}{@{}lccccccc@{}}
\toprule
\textbf{System} & H@1 & H@3 & H@5 & H@10 & H@20 & R@50 & MRR@10 \\
\midrule
\multicolumn{8}{@{}l}{\textit{Encoder-only retrieval}} \\
BM25 & .083 & .113 & .143 & .161 & .226 & .286 & .105 \\
Qwen3-Emb-0.6B & .500 & .613 & .673 & .750 & .815 & .875 & .577 \\
Qwen3-Emb-8B & .560 & .637 & .708 & .756 & .827 & .893 & .618 \\
SR-Emb-0.6B & .560 & .679 & .720 & .762 & .827 & .881 & .629 \\
\textbf{SR-Emb-8B} & \textbf{.673} & \textbf{.780} & \textbf{.798} & \textbf{.851} & \textbf{.887} & \textbf{.929} & \textbf{.729} \\
\midrule
\multicolumn{8}{@{}l}{\textit{Retrieve-and-rerank pipeline}} \\
Qwen3-Emb-0.6B $\times$ Qwen3-Rank-0.6B & .524 & .690 & .738 & .798 & .833 & -- & .620 \\
Qwen3-Emb-8B $\times$ Qwen3-Rank-8B & .661 & .804 & .845 & .875 & .887 & -- & .738 \\
SR-Emb-0.6B $\times$ Qwen3-Rank-0.6B & .601 & .744 & .780 & .827 & .857 & -- & .682 \\
SR-Emb-0.6B $\times$ SR-Rank-0.6B & .679 & .786 & .821 & .839 & .881 & -- & .736 \\
SR-Emb-8B $\times$ Qwen3-Rank-8B & .726 & .845 & .875 & .911 & .923 & -- & .789 \\
\textbf{SR-Emb-8B $\times$ SR-Rank-8B} & \textbf{.786} & \textbf{.851} & \textbf{.881} & \textbf{.923} & \textbf{.929} & -- & \textbf{.828} \\
\bottomrule
\end{tabular}
\end{adjustbox}
\caption{SkillBench-Supp: Results on Descriptive queries (168; 77K pool). R@50 is only defined for encoder-only retrieval, so pipeline rows are marked with `--'.}
\label{tab:supp_breakdown_desc}
\end{table}

\begin{table}[H]
\centering
\small
\begin{adjustbox}{max width=\textwidth}
\begin{tabular}{@{}lccccccc@{}}
\toprule
\textbf{System} & H@1 & H@3 & H@5 & H@10 & H@20 & R@50 & MRR@10 \\
\midrule
\multicolumn{8}{@{}l}{\textit{Encoder-only retrieval}} \\
BM25 & .034 & .057 & .091 & .102 & .148 & .182 & .055 \\
Qwen3-Emb-0.6B & .386 & .511 & .557 & .625 & .693 & .773 & .460 \\
Qwen3-Emb-8B & .398 & .557 & .614 & .716 & .773 & .818 & .494 \\
SR-Emb-0.6B & .352 & .500 & .568 & .648 & .739 & .784 & .446 \\
\textbf{SR-Emb-8B} & \textbf{.477} & \textbf{.648} & \textbf{.693} & \textbf{.773} & \textbf{.841} & \textbf{.898} & \textbf{.573} \\
\midrule
\multicolumn{8}{@{}l}{\textit{Retrieve-and-rerank pipeline}} \\
Qwen3-Emb-0.6B $\times$ Qwen3-Rank-0.6B & .500 & .614 & .648 & .750 & .773 & -- & .570 \\
Qwen3-Emb-8B $\times$ Qwen3-Rank-8B & .591 & .682 & .716 & .807 & .818 & -- & .647 \\
SR-Emb-0.6B $\times$ Qwen3-Rank-0.6B & .523 & .636 & .727 & .761 & .773 & -- & .595 \\
SR-Emb-0.6B $\times$ SR-Rank-0.6B & .568 & .670 & .705 & .750 & .761 & -- & .627 \\
SR-Emb-8B $\times$ Qwen3-Rank-8B & .648 & .761 & .818 & .864 & .898 & -- & .717 \\
\textbf{SR-Emb-8B $\times$ SR-Rank-8B} & \textbf{.625} & \textbf{.784} & \textbf{.818} & \textbf{.841} & \textbf{.875} & -- & \textbf{.701} \\
\bottomrule
\end{tabular}
\end{adjustbox}
\caption{SkillBench-Supp: Results on Indirect queries (88; 77K pool). R@50 is only defined for encoder-only retrieval, so pipeline rows are marked with `--'.}
\label{tab:supp_breakdown_ind}
\end{table}